\newif\ifsuppl \suppltrue  
\newif\ifsmallfonts \smallfontstrue  
\newif\ifcitenum \citenumtrue  

\documentclass{article}

\usepackage{microtype}
\usepackage{booktabs} 

\usepackage{silence}
\usepackage{hyperref}

\usepackage[accepted]{icml2020}

\usepackage{enumitem}
\usepackage{placeins}
\usepackage{subcaption}  
\usepackage{xspace}
\usepackage{booktabs} 
\usepackage{amsmath}
\usepackage[most]{tcolorbox}
\usepackage{tcolorbox}
\usepackage[T1]{fontenc}
\usepackage{amsfonts,amssymb}
\usepackage{bigdelim}
\usepackage{tabularx}
\usepackage{bbm}
\usepackage{calc}
\usepackage{footmisc}
\usepackage{mathrsfs}
\usepackage{mathtools}
\usepackage{stfloats}  
\usepackage{setspace}
\usepackage{tikz}
\usepackage{textgreek}
\usetikzlibrary{tikzmark}
\usetikzlibrary{arrows.meta}
\usetikzlibrary{bending}
\usetikzlibrary{calc,positioning,quotes}
\usetikzlibrary{matrix,shapes.arrows}

\newcommand{\citeopen}{\ifcitenum{}[\else{}(\fi{}}
\newcommand{\citeclose}{\ifcitenum{}]\else{})\fi{}}
\newcommand{\citenobrackets}[1]{\citealp{#1}}  
\newcommand{\citeinline}[1]{\citet{#1}}  
\newcommand{\citewithcomments}[3]{
    \citeopen{}%
    #1
    \citenobrackets{#2}
    #3
    \citeclose{}%
}

\newcommand{\createlength}[2]{\newlength{#1}\setlength{#1}{#2}}

\makeatletter
\DeclareRobustCommand\onedot{\futurelet\@let@token\@onedot}
\def\@onedot{\ifx\@let@token.\else.\null\fi\xspace}
\def\eg{\textit{e.g}\onedot}
\def\Eg{\textit{E.g}\onedot}
\def\ie{\textit{i.e}\onedot}

\def\etc{\textit{etc}\onedot}
\def\vs{\textit{vs}\onedot}
\def\wrt{w.r.t\onedot}

\makeatother

\newcommand{\supplsection}[3]{
    \ifsuppl%
        \section{#1}
        \label{#2}
        #3
    \else%
        \par\refstepcounter{section}%
        \sectionmark{#1}%
        \label{#2}
    \fi%
}

\newcommand{\supplheader}{
    \renewcommand{\figurename}{Supplementary Figure}
    \renewcommand{\tablename}{Supplementary Table}
    \renewcommand{\thesection}{S\arabic{section}}
    \renewcommand{\thefigure}{S\arabic{figure}}
    \renewcommand{\thetable}{S\arabic{table}}
    \setcounter{section}{0}
    \setcounter{figure}{0}
    \setcounter{table}{0}
    \ifsuppl%
        \FloatBarrier
        \clearpage
        \twocolumn[
        \icmltitle{\ourtitle}
        \centerline{\textbf{\LARGE Supplementary Material}}
        \vspace{30pt}
        ]
    \fi%
}

\makeatletter
\renewcommand*{\ALG@name}{Method}
\makeatother

\def\codefont{\fontfamily{lmtt}\selectfont}
\newcommand{\textcode}[1]{{\normalfont\codefont #1}}

\newcommand{\prog}[1]{{\textcode{#1}}\xspace}

\newcounter{codelinecounter}[section]
\definecolor{codebackground}{rgb}{1.0,1.0,1.0}
\definecolor{codebackgroundprogress}{rgb}{1.0,0.99,0.98}
\definecolor{codeframe}{rgb}{0.8,0.8,0.8}
\definecolor{codegreen}{rgb}{0,0.4,0}
\definecolor{codeblue}{rgb}{0.25,0.25,0.75}
\definecolor{codegray}{rgb}{0.5,0.5,0.5}
\def\codefontsize{\fontsize{8.5}{9}\selectfont}  
\newcommand{\codeline}[1]{{#1\par}}
\newcommand{\codelinenum}{\stepcounter{codelinecounter} {\color{codegray} \ifnum\value{codelinecounter}<10 0\fi\arabic{codelinecounter}}\ }

\newcommand{\codecomment}[1]{{\color{codegray} \# #1}}

\newcommand{\codedef}[1]{{\color{codegreen} #1}}
\newcommand{\codereturn}[1]{{\color{codegreen} #1}}
\newcommand{\codetab}{~~}
\newcommand{\codeskip}{\smallskip}
\newenvironment{code}[3]{  
    \setcounter{codelinecounter}{0}
    \begin{tcolorbox}[
        width=#1,height=#2,
        valign=center,left=0pt,right=0pt,top=0pt,bottom=0pt,
        colback=#3,colframe=codeframe,boxrule=0.5pt,arc=0pt]
    \codefont
    \codefontsize
}{ 
    \end{tcolorbox}
}

\newcommand{\experimentdetails}[2][Experiment Details]{
    \begin{spacing}{0.85}%
    {\ifsmallfonts\footnotesize\fi\rule{20pt}{0.4pt}\ \textit{#1: #2}%
    \hrulefill
    }%
    \vspace{-\lineskip}%
    \end{spacing}%
    \vspace{\parskip}%
}

\def\amlnoformat{\mbox{AutoML}}
\def\aml{\amlnoformat\xspace}
\def\amlznoformat{\mbox{AutoML}-\mbox{Zero}}
\def\amlz{\amlznoformat\xspace}

\def\setup{\prog{Setup}}
\def\predict{\prog{Predict}}
\def\learn{\prog{Learn}}

\usepackage{todonotes}

\makeatletter
\newcommand\thefontsize{{[[[This font is \f@size pt]]]}}
\makeatother

\usepackage[english]{babel}
\usepackage{blindtext}
\usepackage{lipsum}

\def\ourtitle{\amlz: Evolving Machine Learning Algorithms From Scratch}

\icmltitlerunning{\amlznoformat}

\begin{document}

\twocolumn[
\icmltitle{\ourtitle}



\icmlsetsymbol{equal}{*}
\icmlsetsymbol{senior}{\textdagger}

\begin{icmlauthorlist}
\icmlauthor{Esteban Real}{equal,google}
\icmlauthor{Chen Liang}{equal,google}
\icmlauthor{David R. So}{google}
\icmlauthor{Quoc V. Le}{google}
\end{icmlauthorlist}

\icmlaffiliation{google}{Google Brain/Google Research, Mountain View, CA, USA}

\icmlcorrespondingauthor{Esteban Real}{ereal@\allowbreak google.com}

\icmlkeywords{machine learning, neural networks, evolution, evolutionary algorithms, regularized evolution, program synthesis, architecture search, NAS, neural architecture search, neuro-architecture search, AutoML, AutoML-Zero, algorithm search, meta-learning, genetic algorithms, genetic programming, neuroevolution, neuro-evolution}

\vskip 0.3in
]

\printAffiliationsAndNotice{\icmlEqualContribution} 

\begin{abstract}
Machine learning research has advanced in multiple aspects, including model structures and learning methods. The effort to automate such research, known as \aml, has also made significant progress. However, this progress has largely focused on the architecture of neural networks, where it has relied on sophisticated expert-designed layers as building blocks---or similarly restrictive search spaces. Our goal is to show that \aml can go further: it is possible today to automatically discover complete machine learning algorithms just using basic mathematical operations as building blocks. We demonstrate this by introducing a novel framework that significantly reduces human bias through a generic search space. Despite the vastness of this space, evolutionary search can still discover two-layer neural networks trained by backpropagation. These simple neural networks can then be surpassed by evolving directly on tasks of interest, \eg CIFAR-10 variants, where modern techniques emerge in the top algorithms, such as bilinear interactions, normalized gradients, and weight averaging. Moreover, evolution adapts algorithms to different task types: \eg, dropout-like techniques appear when little data is available. We believe these preliminary successes in discovering machine learning algorithms from scratch indicate a promising new direction for the field.
\end{abstract}

\section{Introduction}
\label{intro_sec}

In recent years, neural networks have reached remarkable performance on key tasks and seen a fast increase in their popularity \citewithcomments{\eg }{he2015delving,silver2016mastering,wu2016google}{}. This success was only possible due to decades of machine learning (ML) research into many aspects of the field, ranging from learning strategies to new architectures \citewithcomments{}{rumelhart1986learning,lecun1995convolutional,hochreiter1997long}{, among many others}. The length and difficulty of ML research prompted a new field, named \textit{\aml}, that aims to automate such progress by spending machine compute time instead of human research time \cite{fahlman1990cascade,hutter2011sequential,finn2017model}. This endeavor has been fruitful but, so far, modern studies have only employed constrained search spaces heavily reliant on human design. A common example is \textit{architecture search}, which typically constrains the space by only employing sophisticated expert-designed layers as building blocks and by respecting the rules of backpropagation \cite{zoph2016neural,real2017large,tan2019mnasnet}. Other \aml studies similarly have found ways to constrain their search spaces to isolated algorithmic aspects, such as the learning rule used during backpropagation \cite{andrychowicz2016learning,ravi2016optimization}, the data augmentation \cite{cubuk2018autoaugment,park2019specaugment} or the intrinsic curiosity reward in reinforcement learning~\cite{alet2019meta}; in these works, all other algorithmic aspects remain hand-designed. This approach may save compute time but has two drawbacks. First, human-designed components bias the search results in favor of human-designed algorithms, possibly reducing the innovation potential of \aml. Innovation is also limited by having fewer options \cite{elsken2019neural}. Indeed, dominant aspects of performance are often left out \cite{yang2019evaluation}. Second, constrained search spaces need to be carefully composed \cite{zoph2017learning,So2019TheET,negrinho2019towards}, thus creating a new burden on researchers and undermining the purported objective of saving their time.

To address this, we propose to automatically search for \textit{whole} ML algorithms using \textit{little} restriction on form and \textit{only} simple mathematical operations as building blocks. We call this approach \textit{\amlz}, following the spirit of previous work which aims to learn with minimal human participation \citewithcomments{\eg }{silver2017mastering}{}. In other words, \amlz aims to search a fine-grained space simultaneously for the model, optimization procedure, initialization, and so on, permitting much less human-design and even allowing the discovery of \mbox{non-neural} network algorithms. To demonstrate that this is possible today, we present an initial solution to this challenge that creates algorithms competitive with backpropagation-trained neural networks.

The genericity of the \amlz space makes it more difficult to search than existing \aml counterparts. Existing \aml search spaces have been constructed to be dense with good solutions, thus deemphasizing the search method itself. For example, comparisons on the same space found that advanced techniques are often only marginally superior to simple random search (RS) \cite{li2019RS,elsken2019neural,negrinho2019towards}. \amlz is different: the space is so generic that it ends up being quite sparse. The framework we propose represents ML algorithms as computer programs comprised of three \emph{component functions}, \setup, \predict, and \learn, that performs initialization, prediction and learning. The instructions in these functions apply basic mathematical operations on a small memory. The operation and memory addresses used by each instruction are free parameters in the search space, as is the size of the component functions. While this reduces expert design, the consequent sparsity means that RS cannot make enough progress; \eg good algorithms to learn even a trivial task can be as rare as $1$ in $10^{12}$. To overcome this difficulty, we use small proxy tasks and migration techniques to build highly-optimized open-source infrastructure capable of searching through 10,000 models/second/cpu core. In particular, we present a variant of functional equivalence checking that applies to ML algorithms. It prevents re-evaluating algorithms that have already been seen, even if they have different implementations, and results in a 4x speedup. More importantly, for better efficiency, we move away from RS.\footnote{\label{open_source_fn}We open-source our code at \url{https://github.com/google-research/google-research/tree/master/automl_zero\#automl-zero}}

Perhaps surprisingly, evolutionary methods can find solutions in the \amlz search space despite its enormous size and sparsity. By randomly modifying the programs and periodically selecting the best performing ones on given tasks/datasets, we discover reasonable algorithms. We will first show that starting from empty programs and using data labeled by ``teacher'' neural networks with random weights, evolution can discover neural networks trained by gradient descent (Section~\ref{difficulty_sec}). Next, we will minimize bias toward known algorithms by switching to binary classification tasks extracted from CIFAR-10 and allowing a larger set of possible operations. The result is evolved models that surpass the performance of a neural network trained with gradient descent by discovering interesting techniques like multiplicative interactions, normalized gradient and weight averaging (Section~\ref{main_result_sec}). Having shown that these ML algorithms are attainable from scratch, we will finally demonstrate that it is also possible to improve an existing algorithm by initializing the population with it. This way, evolution adapts the algorithm to the type of task provided. For example, dropout-like operations emerge when the task needs regularization and learning rate decay appears when the task requires faster convergence (Section~\ref{zoo_sec}). Additionally, we present ablation studies dissecting our method (Section~\ref{discussion_sec}) and baselines at various compute scales for comparisons by future work (Suppl.\ Section~\ref{baselines_sec}).

\renewcommand\labelitemi{\raisebox{1.5pt}{\fontsize{8}{8}$\bullet$}}

In summary, our contributions are:
\begin{itemize}[noitemsep,topsep=-5pt,leftmargin=*,labelsep=4pt]
    \item \amlz, the proposal to automatically search for ML algorithms from scratch with minimal human design;
    \item A novel framework with open-sourced code\footnotemark[\getrefnumber{open_source_fn}] and a search space that combines only basic mathematical operations;
    \item Detailed results to show potential through the discovery of nuanced ML algorithms using evolutionary search.
\end{itemize}

\section{Related Work}

AutoML has utilized a wide array of paradigms, including growing networks neuron-by-neuron \cite{stanley2002evolving}, hyperparameter optimization \cite{snoek2012practical,loshchilov2016cma,jaderberg2017population} and, neural architecture search~\cite{zoph2016neural,real2017large}.  As discussed in Section~\ref{intro_sec}, \aml has targeted many aspects of neural networks individually, using sophisticated coarse-grained building blocks. \citeinline{mei2019atomnas}, on the other hand, perform a fine-grained search over the convolutions of a neural network. Orthogonally, a few studies benefit from extending the search space to two such aspects simultaneously \cite{zela2018towards,miikkulainen2019evolving,noy2019asap}. In our work, we perform a fine-grained search over all aspects of the algorithm. 

An important aspect of an ML algorithm is the optimization of its weights, which has been tackled by \aml in the form of numerically discovered optimizers \cite{chalmers1991evolution,andrychowicz2016learning,vanschoren2018meta}. The output of these methods is a set of coefficients or a neural network that works well but is hard to interpret. These methods are sometimes described as ``learning the learning algorithm''. However, in our work, we understand \textit{algorithm} more broadly, including the structure and initialization of the model, not just the optimizer. Additionally, our algorithm is not discovered numerically but \textit{symbolically}. A symbolically discovered optimizer, like an equation or a computer program, can be easier to interpret or transfer. An early example of a symbolically discovered optimizer is that of \citeinline{bengio1994use}, who optimize a local learning rule for a 4-neuron neural network using genetic programming \cite{holland1975adaptation,forsyth1981beagle,koza1992genetic}. Our search method is similar but represents the program as a sequence of instructions. While they use the basic operations $\{+, -, \times, \div\}$, we allow many more, taking advantage of dense hardware computations. \citet{risi2010indirectly} tackle the discovery of a biologically informed neural network learning rule too, but with a very different encoding. More recently, \citeinline{bello2017neural} also search for a symbolic optimizer, but in a restricted search space of hand-tuned operations (\eg ``apply dropout with 30\% probability'', ``clip at 0.00001'', \etc). Our search space, on the other hand, aims to minimize restrictions and manual design. Unlike these three studies, we do not even assume the existence of a neural network or of gradients.

We note that our work also relates to program synthesis efforts. Early approaches have proposed to search for programs that improve themselves \cite{lenat1983eurisko,schmidhuber1987evolutionary}. We share similar goals in searching for learning algorithms, but focus on common machine learning tasks and have dropped the self-reflexivity requirement. More recently, program synthesis has focused on solving problems like sorting~\cite{graves2014neural}, string manipulation~\cite{gulwani2017program,balog2017deepcoder}, or structured data QA~\cite{Liang2016NeuralSM}. Unlike these studies, we focus on synthesizing programs that solve the problem of \textit{doing} ML. 

Suppl.\ Section~\ref{additional_related_work} contains additional related work.

\section{Methods}
\label{methods_sec}

\amlz concerns the automatic discovery of algorithms that perform well on a given set of ML tasks $\mathcal{T}$. First, \emph{search experiments} explore a very large space of algorithms $\mathcal{A}$ for an optimal and generalizable $a^* \in \mathcal{A}$. 
The quality of the algorithms is measured on a subset $\mathcal{T}_{search} \subset \mathcal{T}$, with each search experiment producing a candidate algorithm.
In this work, we apply random search as a baseline and evolutionary search as the main search method due to their simplicity and scalability. 
Once the search experiments are done, we select the best candidate by measuring their performances on another subset of tasks $\mathcal{T}_{select} \subset \mathcal{T}$ (\mbox{analo}gous to standard ML model selection with a validation set). Unless otherwise stated, we use binary classification tasks extracted from CIFAR-10, a collection of tiny images each labeled with object classes \cite{krizhevsky2009learning}, and we calculate the average accuracy across a set of tasks to measure the quality of each algorithm. To lower compute costs and achieve higher throughput, we create small proxy tasks for $\mathcal{T}_{search}$ and $\mathcal{T}_{select}$ by using one random matrix for each task to project the input features to lower dimensionality. The projected dimensionality is $8 \leq F \leq 256$. Finally, we compare the best algorithm's performance against hand-designed baselines on the CIFAR-10 data in the original dimensionality (3072), holding out the CIFAR-10 test set for the final evaluation. To make sure the improvement is not specific to CIFAR-10, we further show the gain generalizes to other datasets: SVHN~\cite{netzer2011reading}, ImageNet~\cite{chrabaszcz2017downsampled}, and Fashion MNIST~\cite{xiao2017fashion}. The \textit{Experiment Details} paragraphs in Section~\ref{results_sec} contain the specifics of the tasks. We now describe the search space and search method with sufficient detail to understand the results. For reproducibility, we provide the minutiae in the Supplement and the open-sourced code.

\subsection{Search Space}
\label{methods_search_space_sec}

We represent algorithms as computer programs that act on a small virtual memory with separate address spaces for scalar, vector and matrix variables (\eg\ \textcode{s1}, \textcode{v1}, \textcode{m1}), all of which are floating-point and share the dimensionality of the task's input features ($F$). Programs are sequences of instructions. Each instruction has an operation---or \textit{op}---that determines its function (\eg ``multiply a scalar with a vector''). To avoid biasing the choice of ops, we use a simple criterion: those that are typically learned by high-school level. We purposefully exclude machine learning concepts, matrix decompositions, and derivatives. Instructions have op-specific arguments too. These are typically addresses in the memory (\eg ``read the inputs from scalar address 0 and vector address 3; write the output to vector address 2''). Some ops also require real-valued constants (\eg $\mu$ and $\sigma$ for a random Gaussian sampling op), which are searched for as well. Suppl.\  Section~\ref{search_space_additional_details} contains the full list of 65 ops.

\begin{figure}[!h]
    \begin{code}{0.48\textwidth}{3.35in}{codebackground}  
        \codeline{\codecomment{(Setup, Predict, Learn) = input ML algorithm.}}
        \codeline{\codecomment{Dtrain / Dvalid = training / validation set.}}
        \codeline{\codecomment{sX/vX/mX: scalar/vector/matrix var at address X.}}
        \codeline{\codedef{def} Evaluate(Setup, Predict, Learn, Dtrain, Dvalid):}
        \codeline{\codetab \codecomment{Zero-initialize all the variables (sX/vX/mX).}}
        \codeline{\codetab initialize\_memory()}
        \codeskip
        \codeline{\codetab Setup() \codecomment{Execute setup instructions.}}
        \codeskip
        \codeline{\codetab for (x, y) in Dtrain:}
        \codeline{\codetab \codetab     v0 = x \codecomment{x will now be accessible to Predict.}}
        \codeline{\codetab \codetab     Predict() \codecomment{Execute prediction instructions.}}
        \codeline{\codetab \codetab     \codecomment{s1 will now be used as the prediction.}}
        \codeline{\codetab \codetab     s1 = Normalize(s1)  \codecomment{Normalize the prediction.}}
        \codeline{\codetab \codetab     s0 = y \codecomment{y will now be accessible to Learn.}}
        \codeline{\codetab \codetab     Learn() \codecomment{Execute learning instructions.}}
        \codeskip
        \codeline{\codetab sum\_loss = 0.0}
        \codeline{\codetab for (x, y) in Dvalid:}
        \codeline{\codetab \codetab     v0 = x}
        \codeline{\codetab \codetab     Predict() \codecomment{Only Predict(), not Learn().}}
        \codeline{\codetab \codetab     s1 = Normalize(s1)}
        \codeline{\codetab \codetab     sum\_loss += Loss(y, s1)}
        \codeskip
        \codeline{\codetab mean\_loss = sum\_loss / len(Dvalid)}
        \codeline{\codetab \codecomment{Use validation loss to evaluate the algorithm.}}
        \codeline{\codetab \codereturn{return} mean\_loss}
    \end{code}
\caption{Algorithm evaluation on one task. We represent an algorithm as a program with three component functions (\setup, \predict, \learn). These are evaluated by the pseudo-code above, producing a mean loss for each task. The search method then uses the median across tasks as an indication of the algorithm's quality.}
\label{algorithm_evaluation_fig}
\end{figure}

Inspired by supervised learning work, we represent an algorithm as a program with three \emph{component functions} that we call \setup, \predict, and \learn (\eg Figure~\ref{nn_fig}). The algorithm is evaluated as in Fig~\ref{algorithm_evaluation_fig}. There, the two \mbox{for-loops} implement the \textit{training} and \textit{validation phases}, processing the task's examples one-at-a-time for simplicity. The training phase alternates \predict and \learn executions. Note that \predict just takes in the features of an example (\ie\ \textcode{x})---its label (\ie\ \textcode{y}) is only seen by \learn afterward.

Then, the validation loop executes \predict over the validation examples. After each \predict execution, \textit{whatever} value is in scalar address 1 (\ie{}~\textcode{s1}) is considered the prediction---\predict has no restrictions on what it can write there. For classification tasks, this prediction in $(-\infty,\infty)$ is normalized to a probability in $(0, 1)$ through a sigmoid (binary classification) or a softmax (multi-class). This is implemented as the \textcode{s1 = Normalize(s1)} instruction.
The virtual memory is zero-initialized and persistent, and shared globally throughout the whole evaluation. This way, \setup can initialize memory variables (\eg the weights), \learn can adjust them during training, and \predict can use them. This procedure yields an accuracy for each task. The median across $D$ tasks is used as a measure of the algorithm's quality by the search method.

\subsection {Search Method}
\label{methods_search_sec}

\begin{figure}[!t]
\sbox1{\scalebox{0.45}{\begin{minipage}{0.89in}
\begin{code}{\textwidth}{1.0in}{codebackground}
\tiny
\codeline{\codedef{def} Setup():}
\codeskip
\codeline{\codedef{def} Predict(v0):}
    \codeline{\codetab s9 = arctan(s2) }
    \codeline{\codetab s9 = mean(v5) }
\codeskip
\codeline{\codedef{def} Learn(v0, s0):}
    \codeline{\codetab v1 = s9 * v1 }
\end{code}
\end{minipage}}}
\sbox2{\scalebox{0.45}{\begin{minipage}{0.89in}
\begin{code}{\textwidth}{1.0in}{codebackground}
\tiny
\codeline{\codedef{def} Setup():}
    \codeline{\codetab  s4 = 0.5}
    \codeline{\codetab  s5 = 0.5}
    \codeline{\codetab  m4 = gauss(0,1)}
\codeskip
\codeline{\codedef{def} Predict(v0):}
    \codeline{\codetab v1 = v0 - v9 }
    \codeline{\codetab m4 = s2 * m4 }
\codeskip
\codeline{\codedef{def} Learn(v0, s0):}
    \codeline{\codetab m2 = m2 + m4) }
\end{code}
\end{minipage}}}
\sbox3{\scalebox{0.45}{\begin{minipage}{0.89in}
\begin{code}{\textwidth}{1.0in}{codebackground}
\tiny
\codeline{\codedef{def} Setup():}
\codeskip
\codeline{\codedef{def} Predict(v0):}
\codeskip
\codeline{\codedef{def} Learn(v0, s0):}
\end{code}
\end{minipage}}}
\sbox4{\scalebox{0.45}{\begin{minipage}{0.89in}
\begin{code}{\textwidth}{1.0in}{codebackground}
\tiny
\codeline{\codedef{def} Setup():}
    \codeline{\codetab  s4 = 0.5}
\codeskip
\codeline{\codedef{def} Predict(v0):}
    \codeline{\codetab v1 = v0 - v9 }
    \codeline{\codetab v5 = v0 + v9 }
    \codeline{\codetab m1 = s2 * m2 }
\codeskip
\codeline{\codedef{def} Learn(v0, s0):}
    \codeline{\codetab s4 = s0 - s1 }
    \codeline{\codetab s3 = abs(s1) }
\end{code}
\end{minipage}}}
\sbox5{\scalebox{0.45}{\begin{minipage}{0.89in}
\begin{code}{\textwidth}{1.0in}{codebackground}
\tiny
\codeline{\codedef{def} Setup():}
\codeskip
\codeline{\codedef{def} Predict(v0):}
\codeskip
\codeline{\codedef{def} Learn(v0, s0):}
    \codeline{\codetab v4 = v2 - v1 }
    \codeline{\codetab s3 = mean(v2) }
    \codeline{\codetab s4 = mean(v1) }
    \codeline{\codetab s3 = s3 + s4) }
\end{code}
\end{minipage}}}
\sbox6{\scalebox{0.45}{\begin{minipage}{0.89in}
\begin{code}{\textwidth}{1.0in}{codebackground}
\tiny
\codeline{\codedef{def} Setup():}
    \codeline{\codetab  s4 = 0.5}
\codeskip
\codeline{\codedef{def} Predict(v0):}
    \codeline{\codetab v1 = v0 - v9 }
    \codeline{\codetab v5 = v0 + v9 }
    \codeline{\codetab m1 = s2 * m2 }
\codeskip
\codeline{\codedef{def} Learn(v0, s0):}
    \codeline{\codetab s3 = abs(s1) }
\end{code}
\end{minipage}}}
\begin{tikzpicture}[remember picture]
\createlength{\tinyalgsep}{0.15in}
\createlength{\tinyalgwidth}{0.4in}
\node[inner sep=0] (p0a1) {\usebox1};
\node[right=\tinyalgsep of p0a1, inner sep=0] (p0a2) {\usebox2};
\node[right=\tinyalgsep of p0a2, inner sep=0] (p0a3) {\usebox3};
\node[right=\tinyalgsep of p0a3, inner sep=0] (p0a4) {\usebox4};
\node[right=\tinyalgsep of p0a4, inner sep=0] (p0a5) {\usebox5};
\node[inner sep=0] (p1a1) {\usebox1};
\node[right=\tinyalgsep of p1a1, inner sep=0] (p1a2) {\usebox2};
\node[right=\tinyalgsep of p1a2, inner sep=0] (p1a3) {\usebox3};
\node[right=\tinyalgsep of p1a3, inner sep=0] (p1a4) {\usebox4};
\node[right=\tinyalgsep of p1a4, inner sep=0] (p1a5) {\usebox5};
\node[left=\tinyalgsep of p1a1] (step1) {\footnotesize{Step 1}};
\draw[-,orange,thick] (p1a1.north west) -- (p1a1.south east);
\draw[-,orange,thick] (p1a1.north east) -- (p1a1.south west);
\node[above=0.05in of p1a1] (oldest) {\footnotesize{oldest}};
\node[above=0.05in of p1a5] (newest) {\footnotesize{newest}};
\draw[Straight Barb-Straight Barb,black] (oldest.east) -- ({newest.west});
\node[below=0.3in of p1a1, inner sep=0] (p2a1) {\usebox2};
\node[right=\tinyalgsep of p2a1, inner sep=0] (p2a2) {\usebox3};
\node[right=\tinyalgsep of p2a2, inner sep=0] (p2a3) {\usebox4};
\node[right=\tinyalgsep of p2a3, inner sep=0] (p2a4) {\usebox5};
\node[left=\tinyalgsep of p2a1] (step2) {\footnotesize{Step 2}};
\draw[-Straight Barb,orange,thick,shorten >=3pt] (p1a2) -- (p2a1);
\draw[-Straight Barb,orange,thick,shorten >=3pt] (p1a3) -- (p2a2);
\draw[-Straight Barb,orange,thick,shorten >=3pt] (p1a4) -- (p2a3);
\draw[-Straight Barb,orange,thick,shorten >=3pt] (p1a5) -- (p2a4);
\draw[orange,thick] (p2a1.north west) rectangle (p2a1.south east);
\draw[orange,thick] (p2a3.north west) rectangle (p2a3.south east);
\draw[orange,thick] (p2a4.north west) rectangle (p2a4.south east);
\node[below=0.0in of p2a3] (best) {{\color{orange} \footnotesize{best}}};
\node[below=0.3in of p2a1, inner sep=0] (p3a1) {\usebox2};
\node[right=\tinyalgsep of p3a1, inner sep=0] (p3a2) {\usebox3};
\node[right=\tinyalgsep of p3a2, inner sep=0] (p3a3) {\usebox4};
\node[right=\tinyalgsep of p3a3, inner sep=0] (p3a4) {\usebox5};
\node[right=\tinyalgsep of p3a4, inner sep=0] (p3a5) {\usebox4};
\node[left=\tinyalgsep of p3a1] (step3) {\footnotesize{Step 3}};
\draw[orange,thick] (p3a3.north west) rectangle (p3a3.south east);
\draw [thick,orange,-{[scale=0.7]Straight Barb}]
    (p3a3.south)
    -- ($(p3a3.south)+(0, -7pt)$)
    -- node[below]{{\color{orange} \footnotesize{copy best}}} ($(p3a5.south)+(0, -7pt)$)
    -- (p3a5.south);
\node[below=0.4in of p3a1, inner sep=0] (p4a1) {\usebox2};
\node[right=\tinyalgsep of p4a1, inner sep=0] (p4a2) {\usebox3};
\node[right=\tinyalgsep of p4a2, inner sep=0] (p4a3) {\usebox4};
\node[right=\tinyalgsep of p4a3, inner sep=0] (p4a4) {\usebox5};
\node[right=\tinyalgsep of p4a4, inner sep=0] (p4a5) {\usebox6};
\node[left=\tinyalgsep of p4a1] (step4) {\footnotesize{Step 4}};
\draw [thick,orange,-{[scale=0.7]Straight Barb}]
    (p4a5) to [out=250,in=290,looseness=3] node[below,shift={(0pt, 0pt)}]{{\color{orange} \footnotesize{mutate}}} (p4a5);
\end{tikzpicture}
\begin{tikzpicture}[remember picture,overlay]
\draw[-Straight Barb,black,rounded corners=6pt]
    ($(step4.south)+(-4pt, 0)$)
    -- ($(step4.south)+(-4pt, -7pt)$)
    -- ($(step4.south)+(-19pt, -7pt)$) 
    -- ($(step1.north)+(-19pt, 7pt)$) 
    -- ($(step1.north)+(-4pt, 7pt)$) 
    to ($(step1.north)+(-4pt, 0)$);
\end{tikzpicture}
\caption{One cycle of the evolutionary method \cite{goldberg1991comparative,real2018regularized}. A population of $P$ algorithms (here, $P\!\!=\!\!5$; laid out from left to right in the order they were discovered) undergoes many cycles like this one. First, we remove the oldest algorithm (step 1). Then, we choose a random subset of size $T$ (here, $T\!\!=\!\!3$) and select the best of them (step 2). The best is copied (step 3) and mutated (step 4).
}
\label{evolutionary_method_fig}
\end{figure}
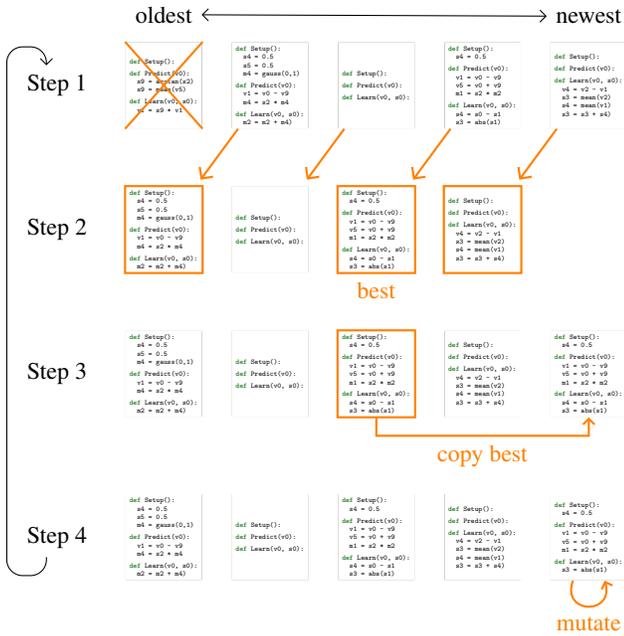

Search experiments must discover algorithms by modifying the instructions in the component functions (\setup, \predict, and \learn; \eg Figure~\ref{nn_fig}). Unless otherwise stated, we use the \textit{regularized evolution} search method because of its simplicity and recent success on architecture search benchmarks \cite{real2018regularized,ying2019bench,So2019TheET}. This method is illustrated in Figure~\ref{evolutionary_method_fig}. It keeps a population of $P$ algorithms, all initially \textit{empty}---\ie none of the three component functions has any instructions/code lines. The population is then improved through cycles. Each cycle picks $T\!\!<\!P$ algorithms at random and selects the best performing one as the \textit{parent}, \ie \emph{tournament selection} \cite{goldberg1991comparative}. This parent is then copied and \textit{mutated} to produce a \textit{child} algorithm that is added to the population, while the oldest algorithm in the population is removed. The mutations that produce the child from the parent must be tailored to the search space; we use a random choice among three types of actions: (i) insert a random instruction or remove an instruction at a random location in a component function, (ii) randomize all the instructions in a component function, or (iii) modify one of the arguments of an instruction by replacing it with a random choice (\eg ``swap the output address'' or ``change the value of a constant''). These are illustrated in Figure~\ref{mutations_fig}.

\newcommand\mutatedtext[1]{\tikz[overlay]\node[fill=orange!20,inner sep=2pt, anchor=text, rectangle, rounded corners=1mm,fill=orange!20] {#1};\phantom{#1}}
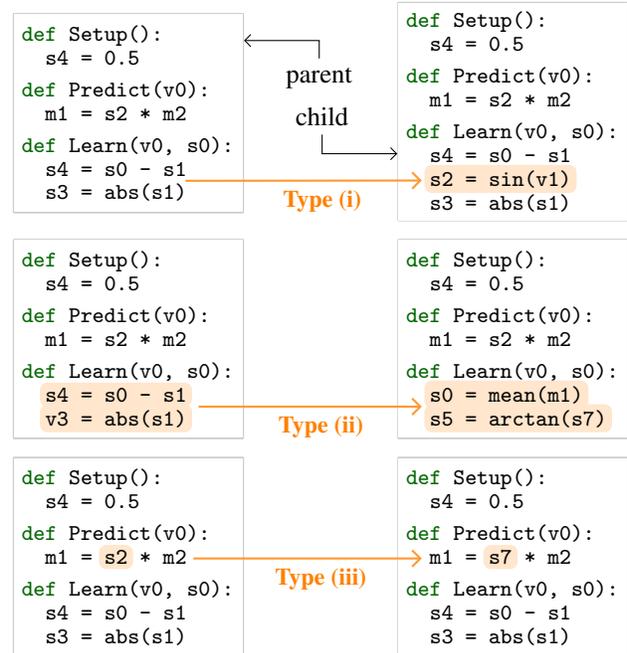
\begin{figure}[!t]
\createlength{\arrowendsep}{3pt}
\sbox1{\begin{subfigure}[t]{1.21in}  
    \begin{code}{\textwidth}{1.05in}{codebackground}  
    \codeline{\codedef{def} Setup():}
        \codeline{\codetab  s4 = 0.5}
    \codeskip
    \codeline{\codedef{def} Predict(v0):}
        \codeline{\codetab m1 = s2 * m2 }
    \codeskip
    \codeline{\codedef{def} Learn(v0, s0):}
        \codeline{\codetab s4 = s0 - s1 \hspace{-5pt}\raisebox{-2pt}{\tikzmark{a}}}
        \codeline{\codetab s3 = abs(s1)}
    \end{code}
\end{subfigure}}
\sbox2{\begin{subfigure}[t]{1.21in}  
    \begin{code}{\textwidth}{1.15in}{codebackground}  
    \codeline{\codedef{def} Setup():}
        \codeline{\codetab  s4 = 0.5}
    \codeskip
    \codeline{\codedef{def} Predict(v0):}
        \codeline{\codetab m1 = s2 * m2 }
    \codeskip
    \codeline{\codedef{def} Learn(v0, s0):}
        \codeline{\codetab s4 = s0 - s1}
        \codeline{\codetab \hspace{-\arrowendsep}\raisebox{2pt}{\tikzmark{b}}\hspace{\arrowendsep}{\mutatedtext{s2 = sin(v1)}} }
        \codeline{\codetab s3 = abs(s1)}
    \end{code}
\end{subfigure}}
\sbox3{\begin{subfigure}[t]{1.21in}  
    \begin{code}{\textwidth}{1.05in}{codebackground}  
    \codeline{\codedef{def} Setup():}
        \codeline{\codetab  s4 = 0.5}
    \codeskip
    \codeline{\codedef{def} Predict(v0):}
        \codeline{\codetab m1 = s2 * m2 }
    \codeskip
    \codeline{\codedef{def} Learn(v0, s0):}
        \codeline{\codetab \mutatedtext{s4 = s0 - s1} \raisebox{-2pt}{\tikzmark{c}}}
        \codeline{\codetab \mutatedtext{v3 = abs(s1)} }
    \end{code}
\end{subfigure}}
\sbox4{\begin{subfigure}[t]{1.21in}  
    \begin{code}{\textwidth}{1.05in}{codebackground}  
    \codeline{\codedef{def} Setup():}
        \codeline{\codetab  s4 = 0.5}
    \codeskip
    \codeline{\codedef{def} Predict(v0):}
        \codeline{\codetab m1 = s2 * m2 }
    \codeskip
    \codeline{\codedef{def} Learn(v0, s0):}
        \codeline{\codetab \hspace{-\arrowendsep}\raisebox{-2pt}{\tikzmark{d}}\hspace{\arrowendsep}{\mutatedtext{s0 = mean(m1)}} }
        \codeline{\codetab \mutatedtext{s5 = arctan(s7)} }
    \end{code}
\end{subfigure}}
\sbox5{\begin{subfigure}[t]{1.21in}  
    \begin{code}{\textwidth}{1.05in}{codebackground}  
    \codeline{\codedef{def} Setup():}
        \codeline{\codetab  s4 = 0.5}
    \codeskip
    \codeline{\codedef{def} Predict(v0):}
        \codeline{\codetab m1 = \mutatedtext{s2} * m2 \hspace{-2pt}\raisebox{2pt}{\tikzmark{e}}}
    \codeskip
    \codeline{\codedef{def} Learn(v0, s0):}
        \codeline{\codetab s4 = s0 - s1 }
        \codeline{\codetab s3 = abs(s1)}
    \end{code}
\end{subfigure}}
\sbox6{\begin{subfigure}[t]{1.21in}  
    \begin{code}{\textwidth}{1.05in}{codebackground}  
    \codeline{\codedef{def} Setup():}
        \codeline{\codetab  s4 = 0.5}
    \codeskip
    \codeline{\codedef{def} Predict(v0):}
        \codeline{\codetab \hspace{-\arrowendsep}\raisebox{2pt}{\tikzmark{f}}\hspace{\arrowendsep}m1 = \mutatedtext{s7} * m2 }
    \codeskip
    \codeline{\codedef{def} Learn(v0, s0):}
        \codeline{\codetab s4 = s0 - s1 }
        \codeline{\codetab s3 = abs(s1)}
    \end{code}
\end{subfigure}}
\setlength{\tabcolsep}{29pt}
\begin{tabular*}{\linewidth}{@{}@{\extracolsep{\fill}}cc}
\raisebox{3pt}{\usebox1\tikzmark{g}} & \tikzmark{h}\usebox2\\[3pt]
\usebox3 & \usebox4\\[3pt]
\usebox5 & \usebox6\\
\end{tabular*}
\begin{tikzpicture}[overlay, remember picture]
\draw [thick,orange!80,-{[scale=1.0]Straight Barb}]
    ({pic cs:a}) to
    node[below,shift={(7pt,0)}] {{\color{orange} \footnotesize{\textbf{Type (i)}}}} ({pic cs:b});
\draw [thick,orange!80,-{[scale=1.0]Straight Barb}]
    ({pic cs:c}) to
    node[below,shift={(4pt,0)}] {{\color{orange} \footnotesize{\textbf{Type (ii)}}}} ({pic cs:d});
\draw [thick,orange!80,-{[scale=1.0]Straight Barb}]
    ({pic cs:e}) to
    node[below,shift={(5pt,0)}] {{\color{orange} \footnotesize{\textbf{Type (iii)}}}} ({pic cs:f});
\node[] (parent) at ([shift={(1, 1.8)}]{pic cs:g}) {parent};
\node[] (child) at ([shift={(-1, 1.4)}]{pic cs:h}) {child};
\draw[-Straight Barb,black] (parent.north) |- ([shift={(0, 2.3)}]{pic cs:g});
\draw[-Straight Barb,black] (child) |- ([shift={(0, 0.9)}]{pic cs:h});
\end{tikzpicture}
\caption{Mutation examples. Parent algorithm is on the left; child on the right. (i) Insert a random instruction (removal also possible). (ii) Randomize a component function. (iii) Modify an argument.}
\label{mutations_fig}
\end{figure}

In order to reach a throughput of $2\textrm{k}$--$10\textrm{k}$ algorithms/second/cpu core, besides the use of small proxy tasks, we apply two additional upgrades: (1) We introduce a version of \textit{functional equivalence checking} (FEC) that detects equivalent supervised ML algorithms, even if they have different implementations, achieving a 4x speedup. To do this, we record the predictions of an algorithm after executing $10$ training and $10$ validation steps on a fixed set of examples. These are then truncated and hashed into a fingerprint for the algorithm to detect duplicates in the population and reuse previous evaluation scores. (2) We add hurdles \cite{So2019TheET} to reach further 5x throughput. In addition to (1) and (2), to attain higher speeds through parallelism, we distribute experiments across worker processes that exchange models through migration \cite{alba2002parallelism}; each process has its own P-sized population and runs on a commodity CPU core. We denote the number of processes by $W$. Typically, $100\!\!<\!\!W\!\!<\!\!1000$ (we indicate the exact numbers with each experiment\footnote{The electricity consumption for our experiments (which were run in 2019) was matched with purchases of renewable energy.}). Workers periodically upload randomly selected algorithms to a central server. The server replies with algorithms randomly sampled across all workers, replacing half the local population (\ie random \emph{migration}). To additionally improve the quality of the search, we allow some workers to search on projected binary MNIST tasks, in addition to projected binary CIFAR-10, to promote diversity (see \eg \cite{wang2019paired}). More details about these techniques can be found in Suppl.\ Section~\ref{detailed_methods_optimization_sec}. Section~\ref{discussion_sec} and Suppl.\ Section~\ref{ablation_comparison_thorough_sec} contain ablation studies showing that all these techniques are beneficial.

For each experimental result, we include an \textit{Experiment Details} paragraph with the exact values for meta-parameters like $P$ and $T$. None of the meta-parameters were tuned in the final set of experiments at full compute scale. Most of them were either decided in smaller experiments (\eg $P$), taken from previous work (\eg $T$), or simply not tuned at all. In some cases, when uncertain about a parameter's appropriate value, we used a range of values instead (\eg ``$100 \leq P \leq 1000$''); different worker processes within the experiment use different values within the range.

\experimentdetails[Details]{Generally, we use $T\!\!=\!\!10$, $100 \leq P \leq 1000$. Each child algorithm is mutated with probability $U\!\!=\!\!0.9$. Run time: $5$ days. Migration rate adjusted so that each worker process has fewer than 1 migration/s and at least 100 migrations throughout the expt. Specifics for each expt.\ in Suppl.\ Section~\ref{detailed_experiment_setups_sec}. Suppl.\  Section~\ref{detailed_methods_optimization_sec} describes additional more general methods minutiae.}

\section{Results}
\label{results_sec}

In the next three sections, we will perform experiments to answer the following three questions, respectively: ``how difficult is searching the \amlz space?'', ``can we use our framework to discover reasonable algorithms with minimal human input?'', and ``can we discover different algorithms by varying the type of task we use during the search experiment?''

\subsection{Finding Simple Neural Nets in a Difficult Space}
\label{difficulty_sec}

We now demonstrate the difficulty of the search space through random search (RS) experiments and we show that, nonetheless, interesting algorithms can be found, especially with evolutionary search. We will explore the benefits of evolution as we vary the task difficulty. We start by searching for algorithms to solve relatively easy problems, such as fitting linear regression data. Note that without the following simplifications, RS would not be able to find solutions.

\experimentdetails{we generate simple regression tasks with $1000$ training and $100$ validation examples with random 8-dim.\ feature vectors $\{x_i\}$ and scalar labels $\{L(x_i)\}$. $L$ is fixed for each task but varies between them. To get affine tasks, $L(x_i)\!=\!u\!\cdot\!x_i\!+\!a$, where $u$ and $a$ are a random vector and scalar. For linear tasks, $a\!\!=\!\!0$. All random numbers were Gaussian ($\mu\!\!=\!\!0$, $\sigma\!\!=\!\!1$). Evaluations use RMS error and the \textcode{Normalize()} instruction in Figure~\ref{algorithm_evaluation_fig} is the identity. We restrict the search space by only using necessary ops and fixing component function lengths to those of known solutions. \Eg, for a linear dataset, \learn has 4 instructions because linear SGD requires 4 instructions. To keep lengths fixed, insert/remove-instruction mutations are not allowed and component functions are initialized randomly. RS generates programs where all instructions are random (see Section~\ref{methods_search_sec}) and selects the best at the end. Evolution expts.\ are small ($W\!\!=\!\!1$; $D\!\!=\!\!3$; $10\textrm{\normalfont k}$ algs./expt.); We repeat expts.\ until statistical significance is achieved. Full configs.\ in Suppl.\ Section~\ref{detailed_experiment_setups_sec}. Note that the restrictions above apply *only* to this section (\ref{difficulty_sec}).}

We quantify a task's difficulty by running a long RS experiment. We count the number of \textit{acceptable algorithms}, \ie those with lower mean RMS error than a hand-designed reference (\eg linear regressor or neural network). The ratio of acceptable algorithms to the total number of algorithms evaluated gives us an \textit{RS success rate}. It can also be interpreted as an estimate of the ``density of acceptable algorithms'' in the search space. We use this density as a measure of problem difficulty. For example, in the linear regression case, we looked for all algorithms that do better than a linear regressor with gradient descent. Even in this trivial task type, we found only 1 acceptable algorithm every $~10^{7}$, so we define $~10^{7}$ to be the difficulty of the linear regression task. We then run the evolution experiments with the same combined total number of evaluations as for RS. We measure the ratio of acceptable algorithms to the total number of algorithms evaluated, to get an \textit{evolution success rate}. However, we only count at most 1 acceptable algorithm from each experiment; this biases the results \emph{against} evolution but is necessary because a single experiment may yield multiple copies of a single acceptable algorithm. Even in the simple case of linear regression, we find that evolution is 5 times more efficient than RS. This stands in contrast to many previous \aml studies, where the solutions are dense enough that RS can be competitive (Section~\ref{intro_sec}).

\begin{figure}[!t]
    \begin{centering}
        \centerline{\includegraphics[width=\columnwidth]{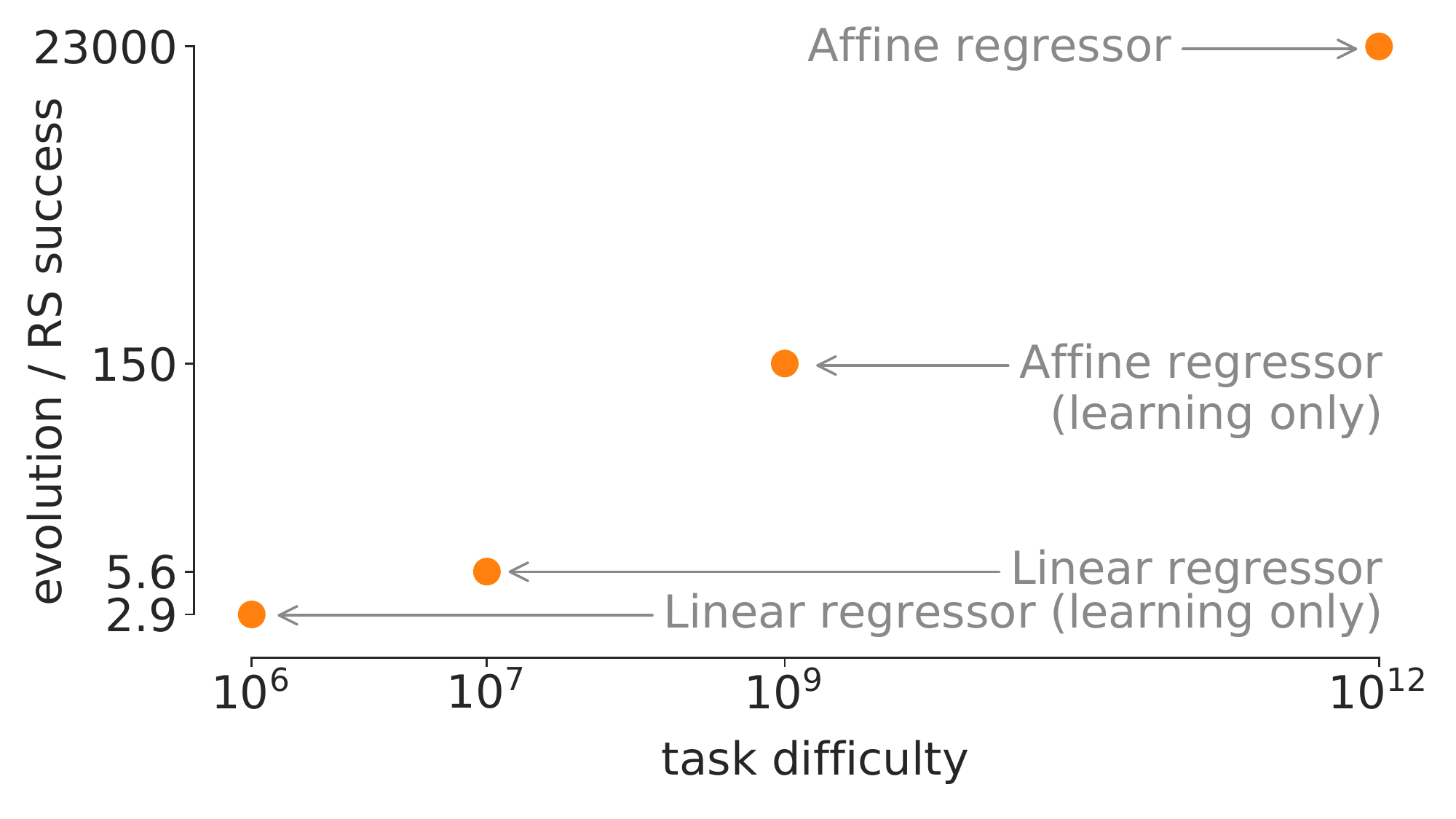}}
        \caption{Relative success rate of evolution and random search (RS). Each point represents a different task type and the x-axis measures its difficulty (defined in the main text). As the task type becomes more difficult, evolution vastly outperforms RS, illustrating the complexity of \amlz when compared to more traditional \aml{} spaces.}
        \label{relative_difficulty_fig}
        \vspace{-2pt}
    \end{centering}
\end{figure}

Figure~\ref{relative_difficulty_fig} summarizes the result of this analysis for 4 task types: the discovery of a full-algorithm/only-the-learning for linear/affine regression data. The \amlz search space is generic but this comes at a cost: even for easy problems, good algorithms are sparse. As the problem becomes more difficult, the solutions become vastly more sparse and evolution greatly outperforms RS.

As soon as we advance to nonlinear data, the gap widens and we can no longer find solutions with RS. To make sure a good solution exists, we generate regression tasks using \emph{teacher} neural networks and then verify that evolution can rediscover the teacher's code.

\experimentdetails{tasks as above but the labeling function is now a teacher network:
$L(x_i)\!=\!u\!\cdot\!\textrm{ReLU}(M\!x_i)$, where
$M$ is a random $8\times8$ matrix, $u$ is a random vector. Number of training examples up to $100\textrm{\normalfont{k}}$. Single expt. Same search space restrictions as above, but now allowing ops used in 2-layer fully connected neural nets. After searching, we select the algorithm with the smallest RMS loss. Full configs.\ in Suppl.\ Section~\ref{detailed_experiment_setups_sec}. Note that the restrictions above apply *only* to this section (\ref{difficulty_sec}).}

When the search method uses only 1 task in $\mathcal{T}_{search}$ (\ie $D\!=\!\!1$), the algorithm evolves the exact prediction function used by the teacher and hard-codes its weights. The results become more surprising as we increase the number of tasks in $\mathcal{T}_{search}$ (\eg to $D\!=\!\!100$), as now the algorithm must find different weights for each task. In this case, evolution not only discovers the forward pass, but also ``invents'' back-propagation code to learn the weights (Figure~\ref{nn_fig}). Despite its difficulty, we conclude that searching the \amlz space seems feasible and we should use evolutionary search instead of RS for more complex tasks.

\begin{figure}[h]
    \begin{code}{0.48\textwidth}{2.75in}{codebackground}
    \codeline{\codecomment{sX/vX/mX = scalar/vector/matrix at address X.}}
    \codeline{\codecomment{``gaussian'' produces Gaussian IID random numbers.}}
    \codeskip
    \codeline{\codedef{def} Setup():}
      \codeline{\codetab \codecomment{Initialize variables.}}
      \codeline{\codetab m1 = gaussian(-1e-10, 9e-09) \codecomment{1st layer weights}}
      \codeline{\codetab s3 = 4.1 \codecomment{Set learning rate}}
      \codeline{\codetab v4 = gaussian(-0.033, 0.01) \codecomment{2nd layer weights}}    \codeskip
    \codeline{\codedef{def} Predict(): \codecomment{v0=features}}
      \codeline{\codetab v6 = dot(m1, v0) \codecomment{Apply 1st layer weights}}
      \codeline{\codetab v7 = maximum(0, v6) \codecomment{Apply ReLU}}
      \codeline{\codetab s1 = dot(v7, v4) \codecomment{Compute prediction}}
    \codeskip
    \codeline{\codedef{def} Learn(): \codecomment{s0=label}}
      \codeline{\codetab v3 = heaviside(v6, 1.0) \codecomment{ReLU gradient}}
      \codeline{\codetab s1 = s0 - s1 \codecomment{Compute error}}
      \codeline{\codetab s2 = s1 * s3 \codecomment{Scale by learning rate}}
      \codeline{\codetab v2 = s2 * v3 \codecomment{Approx.\ 2nd layer weight delta}}
      \codeline{\codetab v3 = v2 * v4 \codecomment{Gradient w.r.t.\ activations}}
      \codeline{\codetab m0 = outer(v3, v0) \codecomment{1st layer weight delta}} 
      \codeline{\codetab m1 = m1 + m0 \codecomment{Update 1st layer weights}}
      \codeline{\codetab v4 = v2 + v4 \codecomment{Update 2nd layer weights}}
    \end{code}
\caption{Rediscovered neural network algorithm. It implements backpropagation by gradient descent. Comments added manually.}
\label{nn_fig}
\end{figure}

\subsection{Searching with Minimal Human Input}
\label{main_result_sec}

Teacher datasets and carefully chosen ops bias the results in favor of known algorithms, so in this section we replace them with more generic options. We now search among a long list of ops selected based on the simplicity criterion described in Section~\ref{methods_search_space_sec}. The increase in ops makes the search more difficult but allows the discovery of solutions other than neural networks. For more realistic datasets, we use binary classification tasks extracted from CIFAR-10 and MNIST.

\begin{figure*}[!htp]
\sbox0{
    \includegraphics[width=\textwidth, trim=5 0 15 0, clip]{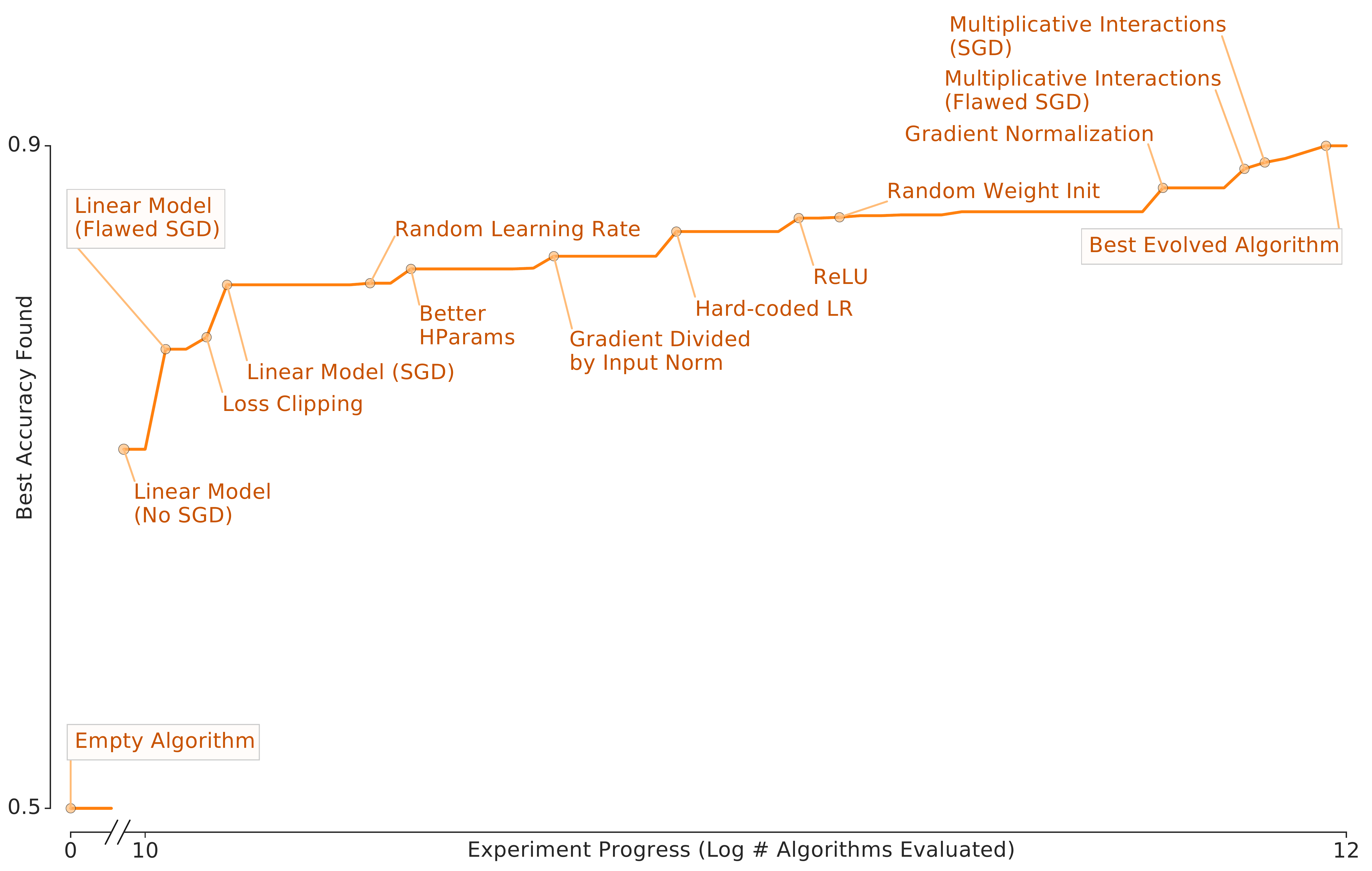}%
}
\sbox1{
    \begin{minipage}{1.2in}
    \begin{code}{\textwidth}{0.88in}{codebackgroundprogress}
    \codeline{\codedef{def} Setup():}
    \codeskip
    \codeline{\codedef{def} Predict():}
    \codeskip
    \codeline{\codedef{def} Learn():}
    \end{code}
    \end{minipage}%
}
\sbox2{
\begin{minipage}{4in}
    \begin{code}{\textwidth}{0.9in}{codebackgroundprogress}
    \begin{minipage}{1.95in}
        \codeline{\codedef{def} Setup():}
        \codeline{\codetab \codecomment{Init weights}}
        \codeline{\codetab v1 = gaussian(0.0, 0.01)}  
        \codeline{\codetab s2 = -1.3}
        \codeskip
        \codeline{\codedef{def} Predict():\ \codecomment{v0=features}}
        \codeline{\codetab s1 = dot(v0, v1) \codecomment{Prediction}}
        \codeskip
    \end{minipage}%
    \begin{minipage}{2in}
        \codeline{\codedef{def} Learn():\ \codecomment{s0=label}}
        \codeline{\codetab s3 = s1 / s2 \codecomment{Scale predict.}}  
        \codeline{\codetab s1 = s0 + s3 \codecomment{Compute error}}
        \codeline{\codetab v2 = s1 * v0 \codecomment{Gradient}}
        \codeline{\codetab v1 = v1 + v2 \codecomment{Update weights}}
        \codeline{\ }
    \end{minipage}%
    \end{code}
    \end{minipage}%
}
\sbox3{
    \begin{minipage}{2.24in}  
    \begin{code}{\textwidth}{2.74in}{codebackgroundprogress}
    \codeline{\codedef{def} Setup():}
    \codeline{\codetab s4 = 1.8e-3 \codecomment{Learning rate}}
    \codeskip
    \codeline{\codedef{def} Predict():\ \codecomment{v0=features}}
        \codeline{\codetab v2 = v0 + v1 \codecomment{Add noise}}
        \codeline{\codetab v3 = v0 - v1 \codecomment{Subtract noise}}
        \codeline{\codetab v4 = dot(m0, v2) \codecomment{Linear}}
        \codeline{\codetab s1 = dot(v3, v4) \codecomment{Mult.interac.}}
        \codeline{\codetab m0 = s2 * m2 \codecomment{Copy weights}}
    \codeskip
    \codeline{\codedef{def} Learn():\ \codecomment{s0=label}}
        \codeline{\codetab s3 = s0 - s1 \codecomment{Compute error}}
        \codeline{\codetab m0 = outer(v3, v0) \codecomment{Approx grad}}
        \codeline{\codetab s2 = norm(m0) \codecomment{Approx grad norm}}
        \codeline{\codetab s5 = s3 / s2 \codecomment{Normalized error}}
        \codeline{\codetab v5 = s5 * v3}
        \codeline{\codetab m0 = outer(v5, v2) \codecomment{Grad}}  
        \codeline{\codetab m1 = m1 + m0 \codecomment{Update weights}}
        \codeline{\codetab m2 = m2 + m1 \codecomment{Accumulate wghts.}}
        \codeline{\codetab m0 = s4 * m1}
        \codeline{\codetab \codecomment{Generate noise}}
        \codeline{\codetab v1 = uniform(2.4e-3, 0.67)}
    \end{code}
    \end{minipage}%
}
\def\diagramextrawidth{35pt}
\sbox4{
    \begin{minipage}[c]{2in+\diagramextrawidth}
    \centering
    \begin{tcolorbox}[
        width=2in+\diagramextrawidth,height=2in,
        valign=center,left=0pt,right=0pt,top=0pt,bottom=0pt,
        colback=codebackgroundprogress,colframe=codeframe,boxrule=0.5pt,arc=0pt]
    \includegraphics[width=1.9in + \diagramextrawidth, trim=7 40 223 5, clip]{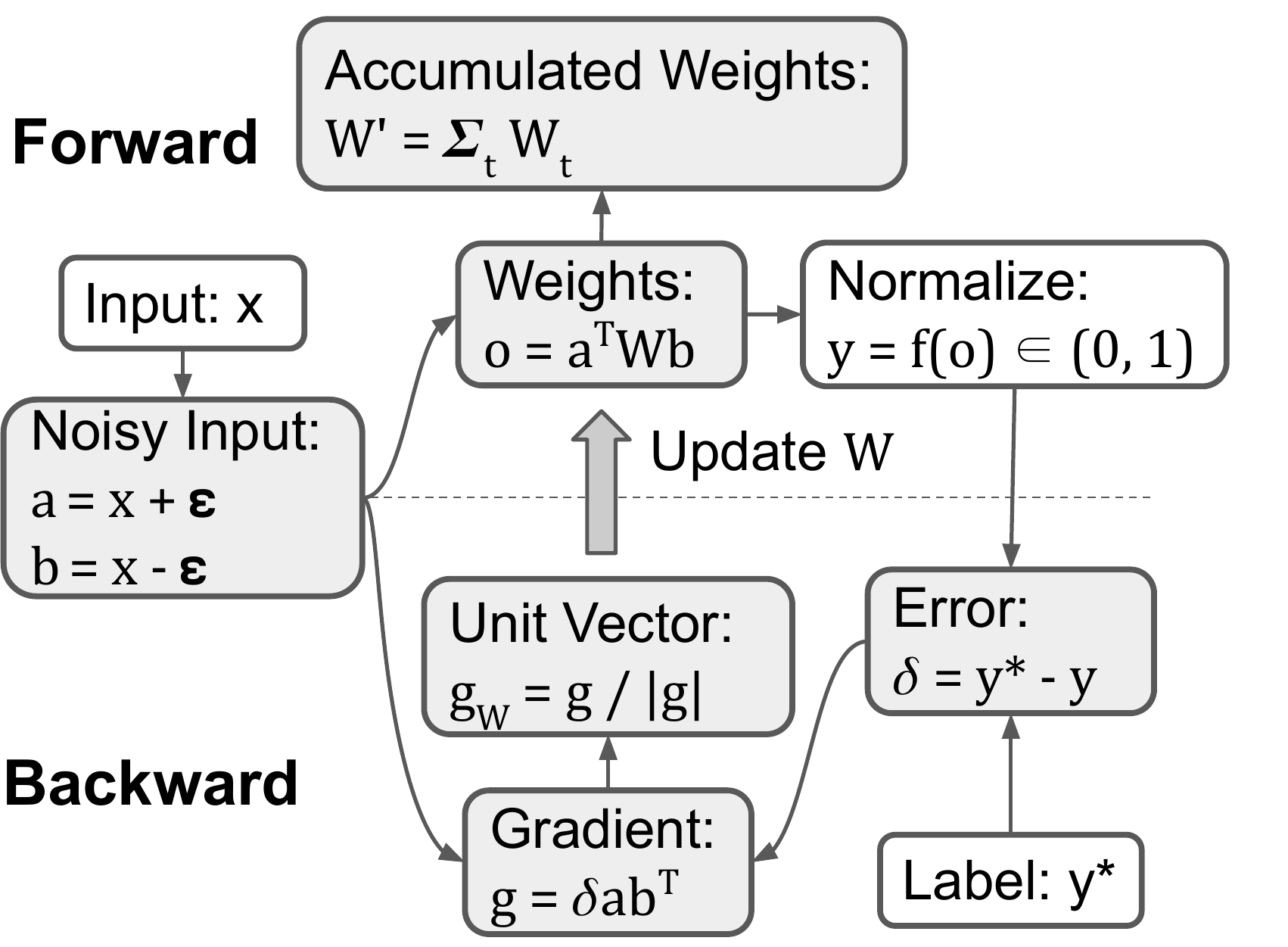}
    \end{tcolorbox}
    \end{minipage}
}
\sbox5{
    \begin{tcolorbox}[
        width=68.55pt,height=3pt,
        valign=center,left=0pt,right=0pt,top=0pt,bottom=0pt,
        colback=codebackgroundprogress,colframe=codebackgroundprogress,boxrule=0.5pt,arc=0pt]
    \end{tcolorbox}
}
\sbox6{
    \begin{tcolorbox}[
        width=56.4pt,height=3pt,
        valign=center,left=0pt,right=0pt,top=0pt,bottom=0pt,
        colback=codebackgroundprogress,colframe=codebackgroundprogress,boxrule=0.5pt,arc=0pt]
    \end{tcolorbox}
}
\sbox8{
    \begin{tcolorbox}[
        width=3pt,height=143.4pt,
        valign=center,left=0pt,right=0pt,top=0pt,bottom=0pt,
        colback=codebackgroundprogress,colframe=codebackgroundprogress,boxrule=0.5pt,arc=0pt]
    \end{tcolorbox}
}
\sbox9{
    \begin{tcolorbox}[
        width=93.4pt,height=3pt,
        valign=center,left=0pt,right=0pt,top=0pt,bottom=0pt,
        colback=codebackgroundprogress,colframe=codebackgroundprogress,boxrule=0.5pt,arc=0pt]
    \end{tcolorbox}
}
%
\createlength{\diagramshift}{92.9pt - \diagramextrawidth}  
\hspace*{-6pt}
\usebox0
\raisebox{81.8pt}{\hspace*{-468.1pt}\usebox1}\hspace*{0.2in}
\raisebox{51pt}{\hspace*{-103.1pt}\usebox5}
\raisebox{274pt}{\hspace*{-74.175pt}\usebox2}
\raisebox{243pt}{\hspace*{-291.05pt}\usebox6}
\raisebox{92.3pt}{\hspace*{\diagramshift}\usebox4}
\raisebox{119pt}{\hspace*{-6pt}\usebox3}
\raisebox{23pt}{\hspace*{-166pt}\usebox8}
\raisebox{218pt}{\hspace*{61.8pt}\usebox9}
\caption{Progress of one evolution experiment on projected binary CIFAR-10. Callouts indicate some beneficial discoveries. We also print the code for the initial, an intermediate, and the final algorithm. The last is explained in the flow diagram. It outperforms a simple fully connected neural network on held-out test data and transfers to features 10x its size. Code notation is the same as in Figure~\ref{nn_fig}. The x-axis gap is due to infrequent recording due to disk throughput limitations.}
\label{experiment_progress_fig}
\end{figure*}

\experimentdetails{We extract tasks from the CIFAR-10 and MNIST training sets; each of the datasets are searched on by half of the processes. For both datasets, the 45 pairs of the 10 classes yield tasks with $8000$ train\,/\,$2000$ valid examples. $36$ pairs are randomly selected to constitute $\mathcal{T}_{search}$, \ie search tasks; $9$ pairs are held out for $\mathcal{T}_{select}$, \i.e tasks for model selection. The CIFAR-10 test set is reserved for final evaluation to report results. Features are projected to $8 \leq F \leq 256$ dim. Each evaluation is on $1 \leq D \leq 10$ tasks. $W\!\!=\!\!10\textrm{k}$. From now on, we use the full setup described in Section~\ref{methods_search_sec}. In particular, we allow variable component function length. Number of possible ops: $7$/\,$58$/\,$58$ for \setup/\,\predict/\,\learn, resp. Full config.\ in Suppl.\ Section~\ref{detailed_experiment_setups_sec}.}

Figure~\ref{experiment_progress_fig} shows the progress of an experiment. It starts with a population of empty programs and automatically invents improvements, several of which are highlighted in the plot. These intermediate discoveries are stepping stones available to evolution and they explain why evolution outperforms RS in this space. 
Each experiment produces a candidate algorithm using $\mathcal{T}_{search}$. 
We then evaluate these algorithms on unseen pairs of classes ($\mathcal{T}_{select}$) and compare the results to a hand-designed reference, a 2-layer fully connected neural network trained by gradient descent. The candidate algorithms perform better in 13 out of 20 experiments. To make sure the improvement is not specific to the small proxy tasks, we select the best algorithm for a final evaluation on binary classification with the original CIFAR-10 data.

Since we are evaluating on tasks with different dimensionality in the final evaluation, we treat all the constants in the best evolved algorithm as hyperparameters and tune them jointly through RS using the validation set. For comparison, we tune two hand-designed baselines, one linear and one nonlinear, using the same total compute that went into discovering and tuning the evolved algorithm. We finally evaluate them all on unseen CIFAR-10 test data. Evaluating with 5 different random seeds, the best evolved algorithm's accuracy ($84.06\pm0.10\%$) significantly outperforms the linear baseline (logistic regression, $77.65\pm0.22\%$) and the nonlinear baseline (2-layer fully connected neural network, $82.22\pm0.17\%$). This gain also generalizes to binary classification tasks extracted from other datasets: SVHN~\cite{netzer2011reading} ($88.12\%$ for the best evolved algorithm \vs $59.58\%$ for the linear baseline \vs $85.14\%$ for the nonlinear baseline), downsampled ImageNet~\cite{chrabaszcz2017downsampled} ($80.78\%$ \vs $76.44\%$ \vs $78.44\%$), Fashion MNIST~\cite{xiao2017fashion} ($98.60\%$ \vs $97.90\%$ \vs $98.21\%$). This algorithm is limited by our simple search space, which cannot currently represent some techniques that are crucial in state-of-the-art models, like batch normalization or convolution. Nevertheless, the algorithm shows interesting characteristics, which we describe below.

\begin{figure*}[!hb]
\createlength{\firstrowheight}{1.73in}  
\createlength{\secondrowheight}{1in}  
\createlength{\firstcolwidth}{2.578in}
\createlength{\secondcolwidth}{1.597in}
\createlength{\thirdcolwidth}{2.316in}
\createlength{\subcaptionspacing}{-10pt}
\def\rowsep{\medskip}
\def\adaptcodesubref{top}
\def\adaptdiagramsubref{bottom}
\sbox0{\begin{subfigure}[@{}c]{\firstcolwidth}
    \begin{code}{\firstcolwidth}{\firstrowheight}{codebackground}
    \codeline{\codedef{def} Predict():}
        \codeline{\codetab ... \codecomment{Omitted/irrelevant code}}
        \codeline{\codetab \codecomment{v0=features; m1=weight matrix}}
        \codeline{\codetab v6 = dot(m1, v0) \codecomment{Apply weights}}
        \codeline{\codetab \codecomment{Random vector, \textmu=-0.5 and \textsigma=0.41}}
        \codeline{\codetab v8 = gaussian(-0.5, 0.41)}
        \codeline{\codetab v6 = v6 + v8 \codecomment{Add it to activations}}
        \codeline{\codetab v7 = maximum(v9, v6) \codecomment{ReLU, v9$\approx$0}}
        \codeline{\codetab ...}
        \codeline{\ }
        \codeline{\ }
        \codeline{\ }
        \codeline{\ }
    \end{code}
\end{subfigure}}
\sbox1{\begin{subfigure}[@{}c]{\secondcolwidth}
    \begin{code}{\secondcolwidth}{\firstrowheight}{codebackground}
    \codeline{\codedef{def} Setup():}
        \codeline{\codetab \codecomment{LR = learning rate}}
        \codeline{\codetab s2 = 0.37 \codecomment{Init.\ LR}}
        \codeline{\codetab ...}
    \codeskip
    \codeline{\codedef{def} Learn():}
        \codeline{\codetab \codecomment{Decay LR}}
        \codeline{\codetab s2 = arctan(s2)}
        \codeline{\codetab ...}
        \codeline{\ }
        \codeline{\ }
        \codeline{\ }
        \codeline{\ }
        \codeline{\ }
    \end{code}
\end{subfigure}}
\sbox2{\begin{subfigure}[@{}c]{\thirdcolwidth}
    \begin{code}{\thirdcolwidth}{\firstrowheight}{codebackground}
    \codeline{\codedef{def} Learn():}
        \codeline{\codetab s3 = mean(m1) \codecomment{m1 is the weights.}}
        \codeline{\codetab s3 = abs(s3)}
        \codeline{\codetab s3 = sin(s3)}
        \codeline{\codetab \codecomment{From here down, s3 is used as}}
        \codeline{\codetab \codecomment{the learning rate.}}
        \codeline{\codetab ...}
        \codeline{\ }
        \codeline{\ }
        \codeline{\ }
        \codeline{\ }
        \codeline{\ }
    \end{code}
\end{subfigure}}
\sbox3{\begin{subfigure}[@{}c]{\firstcolwidth}
    \centering
    \begin{minipage}[c][\secondrowheight][c]{\firstcolwidth}
        \centering
        \includegraphics[width=0.97\firstcolwidth, trim=1 352 412 1, clip]{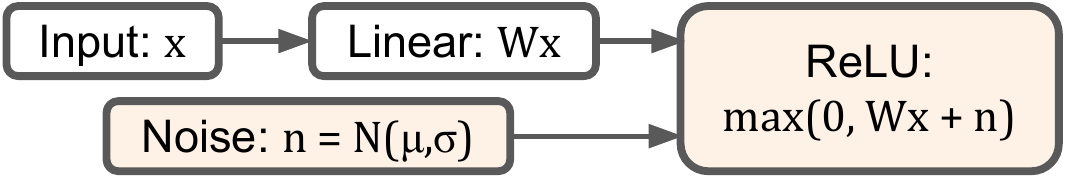}
    \end{minipage}
\end{subfigure}}
\sbox4{\begin{subfigure}[@{}c]{\secondcolwidth}
    \centering
    \begin{minipage}[c][\secondrowheight][c]{\secondcolwidth}
        \centering
        \includegraphics[width=0.95\secondcolwidth, trim=10 5 10 10, clip]{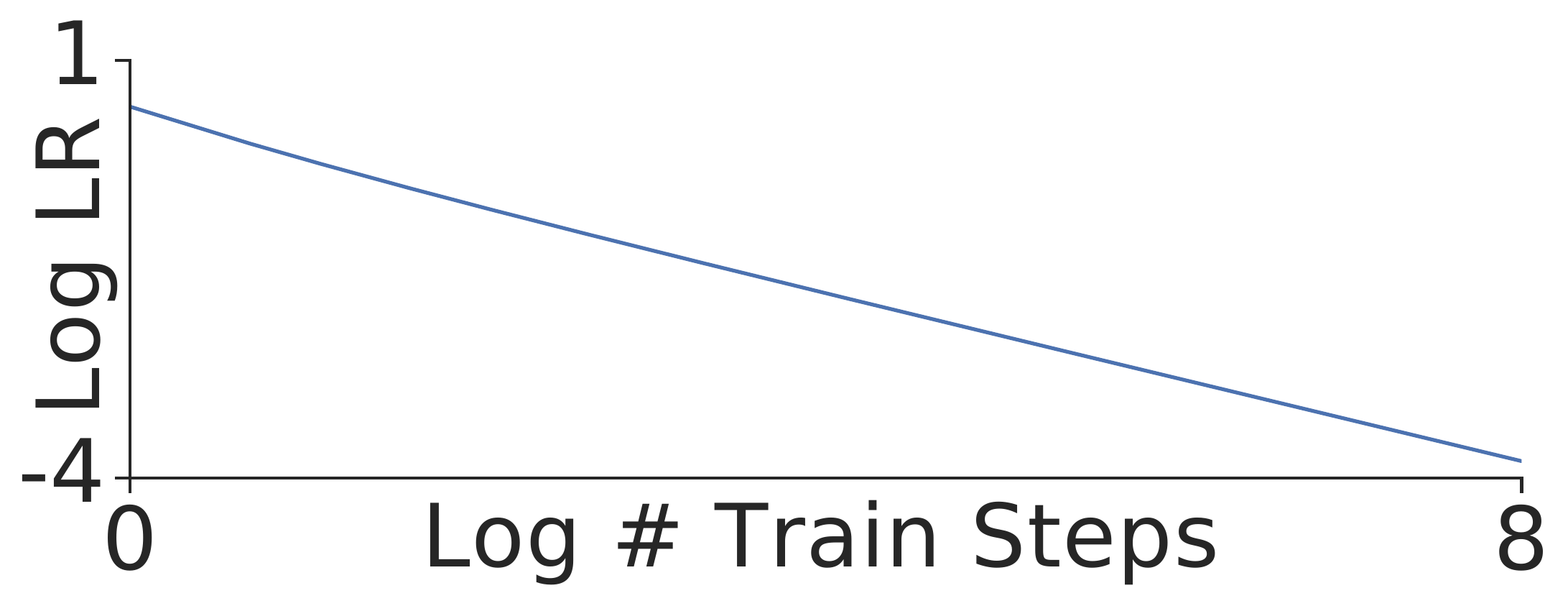}
    \end{minipage}
\end{subfigure}}
\sbox5{\begin{subfigure}[@{}c]{\thirdcolwidth}
    \centering
    \begin{minipage}[c][\secondrowheight][c]{\thirdcolwidth}
        \centering
        \includegraphics[width=0.97\thirdcolwidth, trim=28 355 437 0, clip]{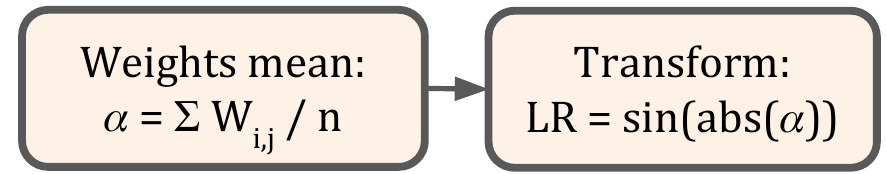}
    \end{minipage}
\end{subfigure}}
\setlength{\tabcolsep}{0pt}
\sbox6{\begin{subfigure}[t]{\firstcolwidth}
    \centering
    \usebox0
    \raisebox{-44pt}{\hspace*{-186pt}\usebox3}
    \vspace{-14pt}
    \caption{Adaptation to few examples.}
    \label{adapt_noisy_relu_fig}
\end{subfigure}}
\sbox7{\begin{subfigure}[t]{\secondcolwidth}
    \centering
    \usebox1
    \raisebox{-40pt}{\hspace*{-115pt}\usebox4}
    \vspace{\subcaptionspacing}
    \caption{Adaptation to fast training.}
    \label{adapt_lr_decay_fig}
\end{subfigure}}
\sbox8{\begin{subfigure}[t]{\thirdcolwidth}
    \centering
    \usebox2
    \raisebox{-40pt}{\hspace*{-168pt}\usebox5}
    \vspace{\subcaptionspacing}
    \caption{Adaptation to multiple classes.}
    \label{adapt_norm_trick_fig}
\end{subfigure}}
\begin{tabular*}{\linewidth}{@{}@{\extracolsep{\fill}}ccc}
\usebox6 & \usebox7 & \usebox8\\
\end{tabular*}
\caption{Adaptations to different task types. (\subref{adapt_noisy_relu_fig})~When few examples are available, evolution creates a noisy ReLU. (\subref{adapt_lr_decay_fig})~When fast training is needed, we get a learning rate decay schedule implemented as an iterated arctan map (\adaptcodesubref) that is nearly exponential (\adaptdiagramsubref). (\subref{adapt_norm_trick_fig})~With multiple classes, the mean of the weight matrix is transformed and then used as the learning rate. Same notation as in Figure~\ref{nn_fig}; full algorithms in Suppl.\ Section~\ref{full_evolved_algorithms_sec}.}
\label{adapt_fig}
\end{figure*}

As a case study, we delve into the best algorithm, shown in Figure~\ref{experiment_progress_fig}. The code has been cleaned for readability; we removed and rearranged instructions when this caused no difference in performance (raw code in Suppl.\  Section~\ref{full_evolved_algorithms_sec}). The algorithm has the following notable features, whose usefulness we verified through ablations (more details in Suppl.\ Section ~\ref{interpretation_supp_sec}):
(1) Noise is added to the input, which, we suspect, acts as a regularizer: 
\begin{equation*}
\mathbf{a} = \mathbf{x} + \mathbf{u}; \mathbf{b} = \mathbf{x} - \mathbf{u}; \mathbf{u} \sim \mathbf{U}(\alpha, \beta)
\end{equation*}
where $\mathbf{x}$ is the input, $\mathbf{u}$ is a random vector drawn from a uniform distribution. 
(2) Multiplicative interactions \cite{jayakumar2020multiplicative} emerge in a \emph{bilinear} form:
$\mathbf{o} = \mathbf{a}^\intercal \mathbf{W} \mathbf{b}$, 
where $\mathbf{o}$ is the output, and $\mathbf{W}$ is the weight matrix.
(3) The gradient $\mathbf{g}$ \wrt the weight matrix $\mathbf{W}$ is computed correctly and is then normalized to be a unit vector:
\begin{equation*}
    \mathbf{g_w} = \frac{\mathbf{g}}{|\mathbf{g}|}; \mathbf{g} = \mathbf{\delta} \mathbf{a} \mathbf{b}^\intercal; \, \mathbf{\delta} = \mathbf{y^*} -  \mathbf{y}; 
\end{equation*}
where $\mathbf{\delta}$ is the error, $y$ is the predicted probability, and $y^*$ is the label. Normalizing gradients is a common heuristic in non-convex optimization~\cite{hazan2015beyond,levy2016power}.
(4) The weight matrix $\mathbf{W'}$ used during inference is the accumulation of all the weight matrices $\{\mathbf{W_t}\}$ after each training step $\mathbf{t}$, \ie: $\mathbf{W'} = \sum_t \mathbf{W_t}$.
This is reminiscent of the \emph{averaged perceptron} \cite{collins2002discriminative} and neural network \emph{weight averaging} during training \cite{polyak1992acceleration,goodfellow2016deep}. Unlike these studies, the evolved algorithm \emph{accumulates} instead of averaging, but this difference has no effect when measuring the accuracy of classification tasks (it does not change the prediction).
As in those techniques, different weights are used at training and validation time. The evolved algorithm achieves this by setting the weights $\mathbf{W}$ equal to $\mathbf{W'}$ at the end of the \predict component function and resetting them to $\mathbf{W_t}$ right after that, at the beginning of the \learn component function. This has no effect during training, when \predict and \learn alternate in execution. However, during validation, \learn is never called and \predict is executed repeatedly, causing $\mathbf{W}$ to remain as $\mathbf{W'}$.

In conclusion, even though the task used during search is simple, the results
show that our framework can discover commonly used algorithms from scratch. 

\subsection{Discovering Algorithm Adaptations}
\label{zoo_sec}

In this section, we will show wider applicability by searching on three different task types. Each task type will impose its own challenge (\eg. ``too little data''). We will show that evolution specifically adapts the algorithms to meet the challenges. Since we already reached reasonable models from scratch above, now we save time by simply initializing the populations with the working neural network of Figure~\ref{nn_fig}.

\experimentdetails{The basic expt.\ configuration and datasets (binary CIFAR-10) are as in Section~\ref{main_result_sec}, with the following exceptions: $W\!\!=\!\!1\textrm{k}$; $F\!\!=\!\!16$; $10 \leq D \leq 100$; critical alterations to the data are explained in each task type below. Full configs.\ in Suppl.\ Section~\ref{detailed_experiment_setups_sec}.}

\textbf{Few training examples.} We use only 80 of the training examples and repeat them for 100 epochs. Under these conditions, algorithms evolve an adaptation that augments the data through the injection of noise (Figure~\ref{adapt_noisy_relu_fig}). This is referred to in the literature as a \textit{noisy ReLU} \cite{nair2010rectified,bengio2013estimating} and is reminiscent of Dropout \cite{srivastava2014dropout}. Was this adaptation a result of the small number of examples or did we simply get lucky? To answer this, we perform $30$ repeats of this experiment and of a control experiment. The control has $800$ examples/100 epochs. We find that the noisy ReLU is reproducible and arises preferentially in the case of little data
(expt: $8/30$, control: $0/30$, $p\!<\!0.0005$).

\textbf{Fast training.} Training on $800$ examples/$10$ epochs leads to the repeated emergence of learning-rate decay, a well-known strategy for the timely training of an ML model \cite{bengio2012practical}. An example can be seen in Figure~\ref{adapt_lr_decay_fig}. As a control, we increase the number of epochs to 100. With overwhelming confidence, the decay appears much more often in the cases with fewer training steps (expt: $30/30$, control: $3/30$, $p\!<\!10^{-14}$).

\textbf{Multiple classes.} When we use all 10 classes of the CIFAR-10 dataset, evolved algorithms tend to use the transformed mean of the weight matrix as the learning rate (Figure~\ref{adapt_norm_trick_fig}). (Note that to support multiple classes, labels and outputs are now vectors, not scalars.) While we do not know the reason, the preference is statistically significant (expt: $24/30$, control: $0/30$, $p\!<\!10^{-11}$).

Altogether, these experiments show that the resulting algorithms seem to adapt well to the different types of tasks.

\section{Conclusion and Discussion}
\label{discussion_sec}

In this paper, we proposed an ambitious goal for \aml: the automatic discovery of whole ML algorithms from basic operations with minimal restrictions on form. The objective was to reduce human bias in the search space, in the hope that this will eventually lead to new ML concepts. As a start, we demonstrated the potential of this research direction by constructing a novel framework that represents an ML algorithm as a computer program comprised of three component functions (\setup, \predict, \learn). Starting from empty component functions and using only basic mathematical operations, we evolved neural networks, gradient descent, multiplicative interactions, weight averaging, normalized gradients, and the like. These results are promising, but there is still much work to be done. In the remainder of this section, we motivate future work with concrete observations.

\begin{figure}[ht]

\begin{centering}
\centerline{\includegraphics[width=\columnwidth]{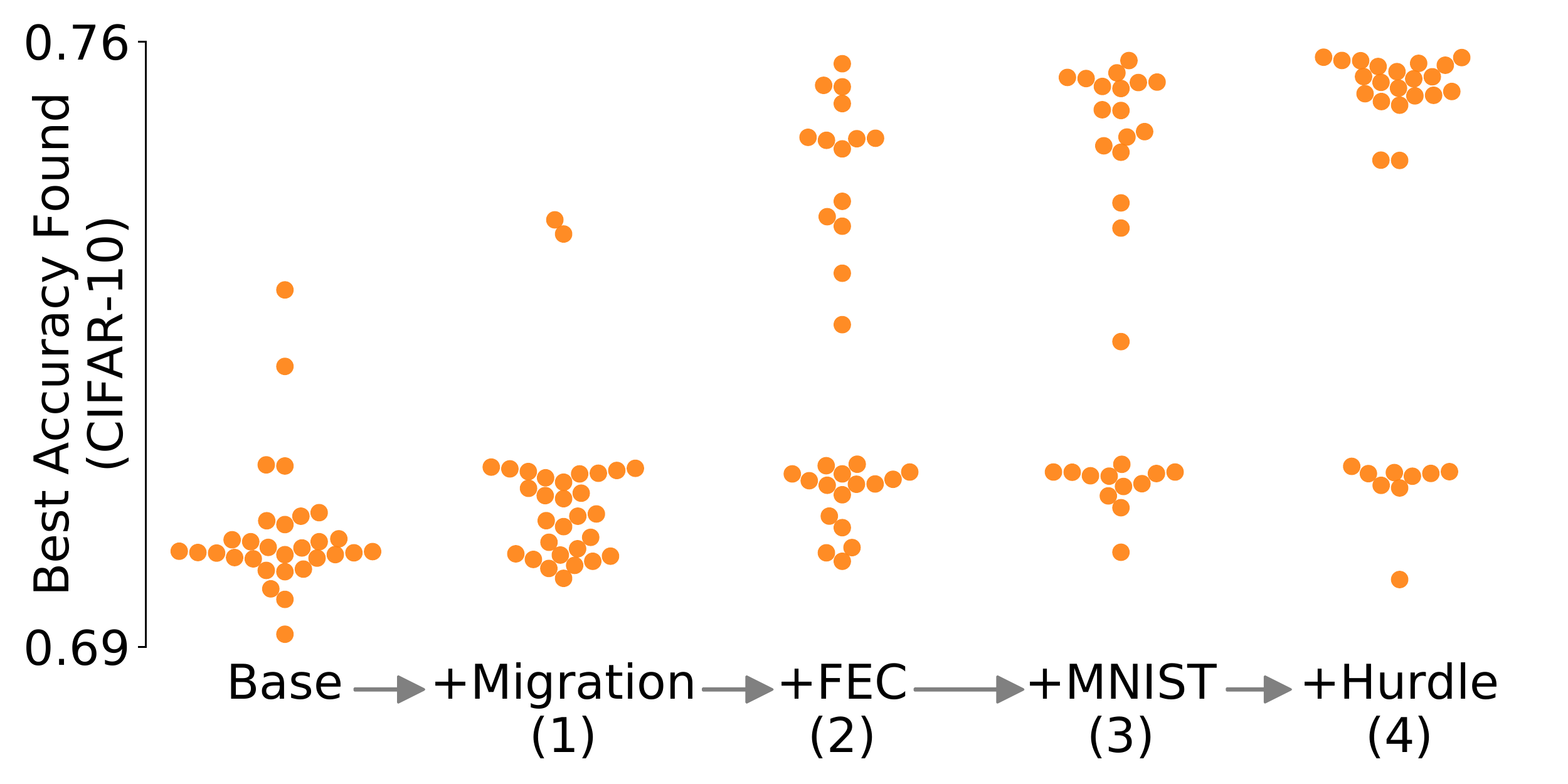}}
\caption{Search method ablation study. From left to right, each column adds an upgrade, as described in the main text.}
\label{ablation_comparison_fig}
\end{centering}
\end{figure}

\textbf{The search method} was not the focus of this study but to reach our results, it helped to (1) add parallelism through migration, (2) use FEC, (3) increase diversity, and (4) apply hurdles, as we detailed in Section~\ref{methods_search_sec}. The effects can be seen in Figure~\ref{ablation_comparison_fig}. Suppl.\ Section~\ref{ablation_comparison_thorough_sec} shows that these improvements work across compute scales (today's high-compute regime is likely to be tomorrow's low-compute regime, so ideas that do not scale with compute will be shorter-lived). Preliminary implementations of crossover and geographic structure did not help in our experiments. The silver lining is that the \amlz search space provides ample room for algorithms to distinguish themselves (\eg Section~\ref{difficulty_sec}), allowing future work to attempt more sophisticated evolutionary approaches, reinforcement learning, Bayesian optimization, and other methods that have helped \aml before.

\textbf{Evaluating evolved algorithms} on new tasks requires hyperparameter tuning, as is common for machine learning algorithms, but without inspection we may not know what each variable means (\eg ``Is \textcode{s7} the learning rate?''). Tuning all constants in the program was insufficient due to \textit{hyperparameter coupling}, where an expression happens to produce a good value for a hyperparameter on a specific set of tasks but won't generalize. For example, evolution may choose \textcode{s7=v2$\cdot$v2} because \textcode{v2$\cdot$v2} coincides with a good value for the hyperparameter \textcode{s7}. We address this through manual decoupling (\eg recognizing the problematic code line and instead setting \textcode{s7} to a constant that can be tuned later). This required time-consuming analysis that could be automated by future work. More details can be found in Suppl.\  Section~\ref{reevaluation_supp_sec}. 

\textbf{Interpreting evolved algorithms} also required effort due to the complexity of their raw code (Suppl.\  Section~\ref{interpretation_supp_sec}). The code was first automatically simplified by removing redundant instructions through static analysis. Then, to decide upon interesting code snippets, Section~\ref{zoo_sec} focused on motifs that reappeared in independent search experiments. Such \emph{convergent evolution} provided a hypothesis that a code section may be beneficial. To verify this hypothesis, we used ablations/\emph{knock-outs} and \emph{knock-ins}, the latter being the insertion of code sections into simpler algorithms to see if they are beneficial there too. This is analogous to homonymous molecular biology techniques used to study gene function. Further work may incorporate other techniques from the natural sciences or machine learning where interpretation of complex systems is key.

\textbf{Search space enhancements} have improved architecture search dramatically. In only two years, for example, comparable experiments went from requiring hundreds of GPUs \cite{zoph2017learning} to only one \cite{liu2018darts}. Similarly, enhancing the search space could bring significant improvements to \amlz. Also note that, despite our best effort to reduce human bias, there is still implicit bias in our current search space that limits the potential to discover certain types of algorithms. For instance, to keep our search space simple, we process one example at a time, so discovering techniques that work on batches of examples (like batch-norm) would require adding loops or higher-order tensors. As another case in point, in the current search space, a multi-layer neural network can only be found by discovering each layer independently; the addition of loops or function calls could make it easier to unlock such deeper structures.

\section*{Author Contributions}
ER and QVL conceived the project; ER led the project; QVL provided advice; ER designed the search space, built the initial framework, and demonstrated plausibility; CL designed proxy tasks, built the evaluation pipeline, and analyzed the algorithms; DRS improved the search method and scaled up the infrastructure; ER, CL, and DRS ran the experiments; ER wrote the paper with contributions from CL; all authors edited the paper and prepared the figures; CL open-sourced the code.

\section*{Acknowledgements}
We would like to thank Samy Bengio, Vincent Vanhoucke, Doug Eck, Charles Sutton, Yanping Huang, Jacques Pienaar, and Jeff Dean for helpful discussions, and especially Gabriel Bender, Hanxiao Liu, Rishabh Singh, Chiyuan Zhang, Hieu Pham, David Dohan and Alok Aggarwal for useful comments on the paper, as well as the larger Google Brain team.


\bibliography{short}
\bibliographystyle{icml2020}


\nocite{fahlman1990cascade,angeline1994evolutionary,yao1999evolving,stanley2002evolving,bergstra2012random,mendoza2016towards,baker2016designing,zoph2016neural,real2017large,xie2017genetic,suganuma2017genetic,liu2017progressive}
\nocite{liu2018darts,elsken2018efficient,cai2018proxylessnas,liu2019auto,ghiasi2019fpn,sun2019evolving,xie2019exploring}
\nocite{elsken2019neural,stanley2019designing,yao2018taking}
\nocite{zoph2017learning,zhong2018practical}
\nocite{mei2019atomnas}
\nocite{snoek2012practical,loshchilov2016cma,jaderberg2017population,li2017hyperband}
\nocite{ramachandran2017searching}
\nocite{kim2015deep}
\nocite{gaier2019weight}
\nocite{cubuk2018autoaugment,park2019specaugment,cubuk2019randaugment}
\nocite{miikkulainen2019evolving}
\nocite{zela2018towards}
\nocite{noy2019asap}
\nocite{chalmers1991evolution}
\nocite{runarsson2000evolution, orchard2016evolution}
\nocite{ravi2016optimization}
\nocite{wichrowska2017learned}
\nocite{metz2019meta}
\nocite{andrychowicz2016learning}
\nocite{li2017learning}
\nocite{risi2010indirectly}
\nocite{vanschoren2018meta}
\nocite{bengio1994use}
\nocite{koza1992genetic}
\nocite{bengio1994use}
\nocite{bello2017neural}
\nocite{bengio1994use}
\nocite{bello2017neural}
\nocite{lenat1983eurisko,schmidhuber1987evolutionary,pitrat1996implementation}
\nocite{schmidhuber2004optimal,graves2014neural,Reed2015NeuralP,Valkov2018HOUDINILL}
\nocite{Polozov2015FlashMetaAF,Parisotto2016NeuroSymbolicPS,Devlin2017RobustFillNP}
\nocite{lake2015human}
\nocite{balog2017deepcoder}
\nocite{Neelakantan2015NeuralPI,Liang2016NeuralSM,Liang2018MemoryAP}
\nocite{chen2017towards}
\nocite{wilson2018evolving}
\nocite{gulwani2017program}
\nocite{real2018regularized}
\nocite{alba2002parallelism}
\nocite{lecun1998mnist}
\nocite{So2019TheET}

\supplheader{}

\supplsection{Additional Related Work}{additional_related_work}{

Because our approach simultaneously searches all the aspects of an ML algorithm, it relates to previous work that targets each aspect individually. As there are many such aspects (\eg architecture, hyperparameters, learning rule), previous work is extensive and impossible to exhaustively list here. Many examples belong within the field of \textit{\aml}. A frequently targeted aspect of the ML algorithm is the structure of the model; this is known as \textit{architecture search}. It has a long history \citewithcomments{}{fahlman1990cascade,angeline1994evolutionary,yao1999evolving,stanley2002evolving,bergstra2012random,mendoza2016towards,baker2016designing,zoph2016neural,real2017large,xie2017genetic,suganuma2017genetic,liu2017progressive}{, and many others} and continues today \citewithcomments{}{liu2018darts,elsken2018efficient,cai2018proxylessnas,liu2019auto,ghiasi2019fpn,sun2019evolving,xie2019exploring}{, and many others}. Reviews provide more thorough background \cite{elsken2019neural,stanley2019designing,yao2018taking}. Recent works have obtained accurate models by constraining the space to only look for the structure of a block that is then stacked to form a neural network. The stacking is fixed and the block is free to combine standard neural network layers into patterns that optimize the accuracy of the model \cite{zoph2017learning,zhong2018practical}. \citeinline{mei2019atomnas} highlight the importance of finer-grained search spaces and take a step in that direction by splitting convolutions into channels that can be handled separately. Other specific architecture aspects have also been targeted, such as the hyperparameters \cite{snoek2012practical,loshchilov2016cma,jaderberg2017population,li2017hyperband}, activation functions \cite{ramachandran2017searching}, a specific layer \cite{kim2015deep}, the full forward pass \cite{gaier2019weight}, the data augmentation \cite{cubuk2018autoaugment,park2019specaugment,cubuk2019randaugment}, \etc. Beyond these narrowly targeted search spaces, \textit{more} inclusive spaces are already demonstrating promise. For example, a few studies have combined two seemingly disparate algorithmic aspects into a single search space: the inner modules and the outer structure \cite{miikkulainen2019evolving}, the architecture and the hyperparameters \cite{zela2018towards}, the layers and the weight pruning \cite{noy2019asap}, and so on. We extend this to all aspects of the algorithm, including the optimization.

An important aspect of an ML algorithm is optimization, which has been tackled by \aml in the form of numerically discovered optimizers. \citeinline{chalmers1991evolution} formalizes the update rule for the weights as \mbox{$w_{i, j} \leftarrow w_{i, j} + F(x_1, x_2, ...)$}, where $x_i$ are local signals and $F$ combines them linearly. The coefficients of the linear combination constitute the search space and are encoded as a bit string that is searched with a genetic algorithm. This is an example of a \textit{numerically learned} update rule: the final result is a set of coefficients that work very well but may not be interpretable. Numerically learned optimizers have improved since then. Studies found that Chalmers' $F$ formula above can be replaced with more advanced structures, such as a second neural network \cite{runarsson2000evolution, orchard2016evolution}, an LSTM \cite{ravi2016optimization}, a hierarchical RNN \cite{wichrowska2017learned}, or even a different LSTM for each weight \cite{metz2019meta}. Numerically or otherwise, some studies center on the method by which the optimizer is learned; it can vary widely from the use of gradient descent \cite{andrychowicz2016learning}, to reinforcement learning \cite{li2017learning}, to evolutionary search with sophisticated developmental encodings \cite{risi2010indirectly}. All these methods are sometimes collectively labeled as \textit{meta-learning} \cite{vanschoren2018meta} or described as ``learning the learning algorithm'', as the optimizer is indeed an algorithm. However, in this work, we understand \textit{algorithm} more broadly and it will include also the structure and the initialization of the model. Additionally, our algorithm is not learned numerically, but discovered \textit{symbolically}. A symbolically discovered optimizer, like an equation or a computer program, can be easier to interpret or transfer.

An early example of a symbolically discovered optimizer is that of \citeinline{bengio1994use}, who represent $F$ as a tree: the leaves are the possible inputs to the optimizer (\ie the $x_i$ above) and the nodes are one of $\{+, -, \times, \div\}$. $F$ is then evolved, making this an example of \textit{genetic programming} \cite{holland1975adaptation,forsyth1981beagle,koza1992genetic}. Our search method is similar to genetic programming but we choose to represent the program as a sequence of instructions---like a programmer would type it---rather than a tree. Another similarity with \citeinline{bengio1994use} is that they also use simple mathematical operations as building blocks. We use many more, however, including vector and matrix instructions that take advantage of dense hardware computations. More recently, \citeinline{bello2017neural} revisited symbolically learned optimizers to apply them to a modern neural network. Their goal was to maximize the final accuracy of their models and so they restrict the search space by allowing hand-tuned operations (\eg ``apply dropout with 30\% probability'', ``clip at 0.00001'', \etc). Our
search space, on the other hand, aims to minimize restrictions and manual design. Both \citeinline{bengio1994use} and \citeinline{bello2017neural} assume the existence of a neural network with a forward pass that computes the activations and a backward pass that provides the weight gradients. Thus, the search process can just focus on discovering how to use these activations and gradients to adjust the network's weights. In contrast, we do not assume the existence of a neural network model or of the gradient. They must therefore be discovered in the same way as the rest of the algorithm.

We note that our work also relates to program synthesis efforts. Early approaches have proposed to search for programs that improve themselves \cite{lenat1983eurisko,schmidhuber1987evolutionary,pitrat1996implementation}. We share similar goals in searching for learning algorithms, but focus on common machine learning tasks and have dropped the self-reflexivity requirement. More recently, program synthesis has focused on solving problems like sorting, addition, counting \cite{schmidhuber2004optimal,graves2014neural,Reed2015NeuralP,Valkov2018HOUDINILL}, string manipulations \cite{Polozov2015FlashMetaAF,Parisotto2016NeuroSymbolicPS,Devlin2017RobustFillNP}, character recognition \cite{lake2015human}, competition-style programming \cite{balog2017deepcoder}, structured data QA \cite{Neelakantan2015NeuralPI,Liang2016NeuralSM,Liang2018MemoryAP}, program parsing \cite{chen2017towards}, and game playing \cite{wilson2018evolving}, to name a few. These studies are increasingly making more \textit{use} of ML to solved the said problems \cite{gulwani2017program}. Unlike these studies, we focus on synthesizing programs that solve the problem of \textit{doing} ML.
}

\supplsection{Search Space Additional Details}{search_space_additional_details}{

Supplementary Table~\ref{ops_table} describes all the ops in our search space. They are ordered to reflect how we chose them: we imagined a typical school curriculum up to---but not including---calculus (see braces to the right of the table).
In particular, there are no derivatives so any gradient computation used for training must be evolved.

\createlength{\rowsep}{3pt}
\def\sepforall{\ \ \forall}
\begin{table*}[!b]
\caption{Ops vocabulary. $s$, $\vec{v}$ and $M$ denote a scalar, vector, and matrix, resp. Early-alphabet letters ($a$, $b$, \etc) denote memory addresses. Mid-alphabet letters (\eg $i$, $j$, \etc) denote vector/matrix indexes (``Index'' column). Greek letters denote constants (``Consts.'' column). $\mathcal{U}(\alpha,\beta)$ denotes a sample from a uniform distribution in $[\alpha,\beta]$. $\mathcal{N}(\mu,\sigma)$ is analogous for a normal distribution with mean $\mu$ and standard deviation $\sigma$. $\mathbbm{1}_X$ is the indicator function for set $X$. Example: ``$M_a^{(i,j)} = \mathcal{U}(\alpha,\beta)$'' describes the operation ``assign to the $i$,$j$-th entry of the matrix at address $a$ a value sampled from a uniform random distribution in $[\alpha,\beta]$''.}
\label{ops_table}
\begin{center}
\begin{small}
\begin{tabular}{p{0.3in}|
                >{\centering\arraybackslash}p{1.3in}|
                >{\centering\arraybackslash}p{0.7in} >{\centering\arraybackslash}p{0.33in}|
                >{\centering\arraybackslash}p{0.5in} >{\centering\arraybackslash}p{0.26in}|
                >{\centering\arraybackslash}p{1.98in}l}
\cmidrule(r{8pt}){1-7}
Op & Code    & \multicolumn{2}{c|}{Input Args} & \multicolumn{2}{c|}{Output Args} & Description   & \\
ID & Example & Addresses & Consts.       & Address    & Index                     & (see caption) & \\
  &          & / types   &               & / type     &                           &               & \\
\cmidrule(r{8pt}){1-7}
OP0 & \textcode{no\_op} & -- & -- & -- & -- & -- & \\[\rowsep]
OP1 & \textcode{s2=s3+s0} & $a$,$b$ / scalars & -- & $c$ / scalar & -- & $s_c = s_a + s_b$ &
\hspace{-15pt} \rdelim\}{7.5}{3pt} \hspace{6pt} \multirow{7.5}{*}{\rotatebox[origin=c]{-90}{Arithmetic}} \\[\rowsep]
OP2 & \textcode{s4=s0-s1} & $a$,$b$ / scalars & -- & $c$ / scalar & -- & $s_c = s_a - s_b$ & \\[\rowsep]
OP3 & \textcode{s8=s5*s5} & $a$,$b$ / scalars & -- & $c$ / scalar & -- & $s_c = s_a \, s_b$ & \\[\rowsep]
OP4 & \textcode{s7=s5/s2} & $a$,$b$ / scalars & -- & $c$ / scalar & -- & $s_c = s_a / s_b$ & \\[\rowsep]
OP5 & \textcode{s8=abs(s0)} & $a$ / scalar & -- & $b$ / scalar & -- & $s_b = |s_a|$ & \\[\rowsep]
OP6 & \textcode{s4=1/s8} & $a$ / scalar & -- & $b$ / scalar & -- & $s_b = 1/s_a$ & \\[\rowsep]
OP7 & \textcode{s5=sin(s4)} & $a$ / scalar & -- & $b$ / scalar & -- & $s_b = \sin(s_a)$ &
\hspace{-15pt} \rdelim\}{7.5}{3pt} \hspace{6pt} \multirow{7.5}{*}{\rotatebox[origin=c]{-90}{Trigonometry}} \\[\rowsep]
OP8& \textcode{s1=cos(s4)} & $a$ / scalar & -- & $b$ / scalar & -- & $s_b = \cos(s_a)$ & \\[\rowsep]
OP9 & \textcode{s3=tan(s3)} & $a$ / scalar & -- & $b$ / scalar & -- & $s_b = \tan(s_a)$ & \\[\rowsep]
OP10 & \textcode{s0=arcsin(s4)} & $a$ / scalar & -- & $b$ / scalar & -- & $s_b = \arcsin(s_a)$ & \\[\rowsep]
OP11 & \textcode{s2=arccos(s0)} & $a$ / scalar & -- & $b$ / scalar & -- & $s_b = \arccos(s_a)$ & \\[\rowsep]
OP12 & \textcode{s4=arctan(s0)} & $a$ / scalar & -- & $b$ / scalar & -- & $s_b = \arctan(s_a)$ & \\[\rowsep]
OP13 & \textcode{s1=exp(s2)} & $a$ / scalar & -- & $b$ / scalar & -- & $s_b = e^{s_a}$ &
\hspace{-15pt} \rdelim\}{6.5}{3pt} \hspace{6pt} \multirow{6.5}{*}{\rotatebox[origin=c]{-90}{Pre-Calculus}} \\[\rowsep]
OP14 & \textcode{s0=log(s3)} & $a$ / scalar & -- & $b$ / scalar & -- & $s_b = \operatorname{log}{s_a}$ & \\[\rowsep]
OP15 & \textcode{s3=heaviside(s0)} & $a$ / scalar & -- & $b$ / scalar & -- & $s_b = \mathbbm{1}_{\mathbb{R}^+}(s_a)$ & \\[\rowsep]
OP16 & \textcode{v2=heaviside(v2)} & $a$ / vector & -- & $b$ / vector & -- & $\vec{v}_b^{\,(i)} = \mathbbm{1}_{\mathbb{R}^+}(\vec{v}_a^{\,(i)}) \sepforall i$ & \\[\rowsep]
OP17 & \textcode{m7=heaviside(m3)} & $a$ / matrix & -- & $b$ / matrix & -- & $M_b^{(i,j)} = \mathbbm{1}_{\mathbb{R}^+}(M_a^{(i,j)}) \sepforall i,j$ & \\[\rowsep]
OP18 & \textcode{v1=s7*v1} & $a$,$b$ / sc,vec & -- & $c$ / vector & -- & $\vec{v}_c = s_a \, \vec{v}_b$ &
\hspace{-15pt} \rdelim\}{8.1}{3pt} \hspace{6pt} \multirow{8.1}{*}{\rotatebox[origin=c]{-90}{Linear Algebra}} \\[\rowsep]
OP19 & \textcode{v1=bcast(s3)} & $a$ / scalar & -- & $b$ / vector & -- & $\vec{v}_b^{\,(i)} = s_a \ \ \sepforall i$ & \\[\rowsep]
OP20 & \textcode{v5=1/v7} & $a$ / vector & -- & $b$ / vector & -- & $\vec{v}_b^{\,(i)} = 1/\vec{v}_a^{\,(i)} \ \ \sepforall i$ & \\[\rowsep]
OP21 & \textcode{s0=norm(v3)} & $a$ / scalar & -- & $b$ / vector & -- & $s_b = |\vec{v}_a|$ & \\[\rowsep]
OP22 & \textcode{v3=abs(v3)} & $a$ / vector & -- & $b$ / vector & -- & $\vec{v}_b^{\,(i)} = |\vec{v}_a^{\,(i)}| \ \ \sepforall i$ & \\[\rowsep]
\multicolumn{7}{c}{\dotfill [Table continues on the next page.] \dotfill}\\
\end{tabular}
\end{small}
\end{center}
\end{table*}

\addtocounter{table}{-1}
\begin{table*}[!t]
\vspace*{20pt}
\caption{Ops vocabulary (continued)}
\begin{center}
\begin{small}
\begin{tabular}{p{0.3in}|
                >{\centering\arraybackslash}p{1.3in}|
                >{\centering\arraybackslash}p{0.7in} >{\centering\arraybackslash}p{0.33in}|
                >{\centering\arraybackslash}p{0.5in} >{\centering\arraybackslash}p{0.26in}|
                >{\centering\arraybackslash}p{1.98in}l}
\cmidrule(r{8pt}){1-7}
Op & Code    & \multicolumn{2}{c|}{Input Args} & \multicolumn{2}{c|}{Output Args} & Description   & \\
ID & Example & Addresses & Consts       & Address    & Index                     & (see caption) & \\
  &          & / types   &              & / type     &                           &               & \\
\cmidrule(r{8pt}){1-7}
OP23 & \textcode{v5=v0+v9} & $a$,$b$ / vectors & -- & $c$ / vector & -- & $\vec{v}_c = \vec{v}_a + \vec{v}_b$ &
\hspace{-15pt} \rdelim\}{29}{3pt} \hspace{6pt} \multirow{29}{*}{\rotatebox[origin=c]{-90}{Linear Algebra}} \\[\rowsep]
OP24 & \textcode{v1=v0-v9} & $a$,$b$ / vectors & -- & $c$ / vector & -- & $\vec{v}_c = \vec{v}_a - \vec{v}_b$ & \\[\rowsep]
OP25 & \textcode{v8=v1*v9} & $a$,$b$ / vectors & -- & $c$ / vector & -- & $\vec{v}_c^{\,(i)} = \vec{v}_a^{\,(i)} \, \vec{v}_b^{\,(i)} \sepforall i$ & \\[\rowsep]
OP26 & \textcode{v9=v8/v2} & $a$,$b$ / vectors & -- & $c$ / vector & -- & $\vec{v}_c^{\,(i)} = \vec{v}_a^{\,(i)} / \vec{v}_b^{\,(i)} \sepforall i$ & \\[\rowsep]
OP27 & \textcode{s6=dot(v1,v5)} & $a$,$b$ / vectors & -- & $c$ / scalar & -- & $s_c=\vec{v}_a^{\,T} \, \vec{v}_b$ & \\[\rowsep]
OP28 & \textcode{m1=outer(v6,v5)} & $a$,$b$ / vectors & -- & $c$ / matrix & -- & $M_c=\vec{v}_a \, \vec{v}_b^{\,T}$ & \\[\rowsep]
OP29 & \textcode{m1=s4*m2} & $a$,$b$ / sc/mat & -- & $c$ / matrix & -- & $M_c=s_a \, M_b$ & \\[\rowsep]
OP30 & \textcode{m3=1/m0} & $a$ / matrix & -- & $b$ / matrix & -- & $M_b^{(i,j)}=1/M_a^{(i,j)} \sepforall i,j$ & \\[\rowsep]
OP31 & \textcode{v6=dot(m1,v0)} & $a$,$b$ / mat/vec & -- & $c$ / vector & -- & $\vec{v}_c = M_a \, \vec{v}_b$ & \\[\rowsep]
OP32 & \textcode{m2=bcast(v0,axis=0)} & $a$ / vector & -- & $b$ / matrix & -- & $M_b^{(i,j)} = \vec{v}_a^{\,(i)} \sepforall i,j$ & \\[\rowsep]
OP33 & \textcode{m2=bcast(v0,axis=1)} & $a$ / vector & -- & $b$ / matrix & -- & $M_b^{(j,i)} = \vec{v}_a^{\,(i)} \sepforall i,j$ & \\[\rowsep]
OP34 & \textcode{s2=norm(m1)} & $a$ / matrix & -- & $b$ / scalar & -- & $s_b = ||M_a||$ & \\[\rowsep]
OP35 & \textcode{v4=norm(m7,axis=0)} & $a$ / matrix & -- & $b$ / vector & -- & $\vec{v}_b^{\,(i)} = |M_a^{(i,\cdot)}| \sepforall i$ & \\[\rowsep]
OP36 & \textcode{v4=norm(m7,axis=1)} & $a$ / matrix & -- & $b$ / vector & -- & $\vec{v}_b^{\,(j)} = |M_a^{(\cdot,j)}| \sepforall j$ & \\[\rowsep]
OP37 & \textcode{m9=transpose(m3)} & $a$ / matrix & -- & $b$ / matrix & -- & $M_b = |M_a^T|$ & \\[\rowsep]
OP38 & \textcode{m1=abs(m8)} & $a$ / matrix & -- & $b$ / matrix & -- & $M_b^{(i,j)} = |M_a^{(i,j)}| \sepforall i,j$ & \\[\rowsep]
OP39 & \textcode{m2=m2+m0} & $a$,$b$ / matrixes & -- & $c$ / matrix & -- & $M_c = M_a + M_b$ & \\[\rowsep]
OP40 & \textcode{m2=m3+m1} & $a$,$b$ / matrixes & -- & $c$ / matrix & -- & $M_c = M_a - M_b$ & \\[\rowsep]
OP41 & \textcode{m3=m2*m3} & $a$,$b$ / matrixes & -- & $c$ / matrix & -- & $M_c^{(i,j)} = M_a^{(i,j)} \, M_b^{(i,j)} \sepforall i,j$ & \\[\rowsep]
OP42 & \textcode{m4=m2/m4} & $a$,$b$ / matrixes & -- & $c$ / matrix & -- & $M_c^{(i,j)} = M_a^{(i,j)} / M_b^{(i,j)} \sepforall i,j$ & \\[\rowsep]
OP43 & \textcode{m5=matmul(m5,m7)} & $a$,$b$ / matrixes & -- & $c$ / matrix & -- & $M_c = M_a \, M_b$ & \\[\rowsep]
OP44 & \textcode{s1=minimum(s2,s3)} & $a$,$b$ / scalars & -- & $c$ / scalar & -- & $s_c = \min(s_a, s_b)$ &
\hspace{-15pt} \rdelim\}{25}{3pt} \hspace{6pt} \multirow{25}{*}{\rotatebox[origin=c]{-90}{Probability and Statistics}} \\[\rowsep]
OP45 & \textcode{v4=minimum(v3,v9)} & $a$,$b$ / vectors & -- & $c$ / vector & -- & $\vec{v}_c^{\,(i)} = \min(\vec{v}_a^{\,(i)}, \vec{v}_b^{\,(i)}) \sepforall i$ & \\[\rowsep]
OP46 & \textcode{m2=minimum(m2,m1)} & $a$,$b$ / matrixes & -- & $c$ / matrix & -- & $M_c^{(i,j)} = \min(M_a^{(i,j)}, M_b^{(i,j)}) \sepforall i,j$ & \\[\rowsep]
OP47 & \textcode{s8=maximum(s3,s0)} & $a$,$b$ / scalars & -- & $c$ / scalar & -- & $s_c = \max(s_a, s_b)$ & \\[\rowsep]
OP48 & \textcode{v7=maximum(v3,v6)} & $a$,$b$ / vectors & -- & $c$ / vector & -- & $\vec{v}_c^{\,(i)} = \max(\vec{v}_a^{\,(i)}, \vec{v}_b^{\,(i)}) \sepforall i$ & \\[\rowsep]
OP49 & \textcode{m7=maximum(m1,m0)} & $a$,$b$ / matrixes & -- & $c$ / matrix & -- & $M_c^{(i,j)} = \max(M_a^{(i,j)}, M_b^{(i,j)}) \sepforall i,j$ & \\[\rowsep]
OP50 & \textcode{s2=mean(v2)} & $a$ / vector & -- & $b$ / scalar & -- & $s_b = \operatorname{mean}(\vec{v}_a)$ & \\[\rowsep]
OP51 & \textcode{s2=mean(m8)} & $a$ / matrix & -- & $b$ / scalar & -- & $s_b = \operatorname{mean}(M_a)$ & \\[\rowsep]
OP52 & \textcode{v1=mean(m2,axis=0)} & $a$ / matrix & -- & $b$ / vector & -- & $\vec{v}_b^{\,(i)} = \operatorname{mean}(M_a^{(i,\cdot)}) \sepforall i$ & \\[\rowsep]
OP53 & \textcode{v3=std(m2,axis=0)} & $a$ / matrix & -- & $b$ / vector & -- & $\vec{v}_b^{\,(i)} = \operatorname{stdev}(M_a^{(i,\cdot)}) \sepforall i$ & \\[\rowsep]
OP54 & \textcode{s3=std(v3)} & $a$ / vector & -- & $b$ / scalar & -- & $s_b = \operatorname{stdev}(\vec{v}_a)$ & \\[\rowsep]
OP55 & \textcode{s4=std(m0)} & $a$ / matrix & -- & $b$ / scalar & -- & $s_b = \operatorname{stdev}(M_a)$ & \\[\rowsep]
OP56 & \textcode{s2=0.1} & -- & $\gamma$ & $a$ / scalar & -- & $s_a = \gamma$ & \\[\rowsep]
OP57 & \textcode{v3[5]=-2.4} & -- & $\gamma$ & $a$ / vector & $i$ & $\vec{v}_a^{\,(i)} = \gamma$ & \\[\rowsep]
OP58 & \textcode{m2[5,1]=-0.03} & -- & $\gamma$ & $a$ / matrix & $i$, $j$ & $M_a^{(i,j)} = \gamma$ & \\[\rowsep]
OP59 & \textcode{s4=uniform(-1,1)} & -- & $\alpha$, $\beta$ & $a$ / scalar & -- & $s_a = \mathcal{U}(\alpha,\beta)$ & \\[\rowsep]
OP60 & \textcode{v1=uniform(0.4,0.8)} & -- & $\alpha$, $\beta$ & $a$ / vector & -- & $\vec{v}_a^{\,(i)} = \mathcal{U}(\alpha,\beta) \sepforall i$ & \\[\rowsep]
\multicolumn{7}{c}{\dotfill [Table continues on the next page.] \dotfill}\\
\end{tabular}
\end{small}
\end{center}
\end{table*}

\addtocounter{table}{-1}
\begin{table*}[!t]
\caption{Ops vocabulary (continued)}
\begin{center}
\begin{small}
\begin{tabular}{p{0.3in}|
                >{\centering\arraybackslash}p{1.3in}|
                >{\centering\arraybackslash}p{0.7in} >{\centering\arraybackslash}p{0.33in}|
                >{\centering\arraybackslash}p{0.5in} >{\centering\arraybackslash}p{0.26in}|
                >{\centering\arraybackslash}p{1.98in}l}
\cmidrule(r{8pt}){1-7}
Op & Code    & \multicolumn{2}{c|}{Input Args} & \multicolumn{2}{c|}{Output Args} & Description   & \\
ID & Example & Addresses & Consts       & Address    & Index                     & (see caption) & \\
  &          & / types   &              & / type     &                           &               & \\
\cmidrule(r{8pt}){1-7}
OP61 & \textcode{m0=uniform(-0.5,0.6)} & -- & $\alpha$, $\beta$ & $a$ / matrix & -- & $M_a^{(i,j)} = \mathcal{U}(\alpha,\beta) \sepforall i,j$ &
\hspace{-15pt} \rdelim\}{5.4}{3pt} \hspace{6pt} \multirow{5.4}{*}{\rotatebox[origin=c]{-90}{Prob.\ and Stats.}} \\[\rowsep]
OP62 & \textcode{s4=gaussian(0.1,0.7)} & -- & $\mu$, $\sigma$ & $a$ / scalar & -- & $s_a = \mathcal{N}(\mu,\sigma)$ & \\[\rowsep]
OP63 & \textcode{v8=gaussian(0.4,1)} & -- & $\mu$, $\sigma$ & $a$ / vector & -- & $\vec{v}_a^{\,(i)} = \mathcal{N}(\mu,\sigma) \sepforall i$ & \\[\rowsep]
OP64 & \textcode{m2=gaussian(-2,1.3)} & -- & $\mu$, $\sigma$ & $a$ / matrix & -- & $M_a^{(i,j)} = \mathcal{N}(\mu,\sigma) \sepforall i,j$ \\[\rowsep]
\cmidrule(r{8pt}){1-7}
\end{tabular}
\end{small}
\end{center}
\end{table*}

}  

\supplsection{Search Method Additional Details}{detailed_methods_optimization_sec}{

The mutations that produce the child from the parent must be tailored to the search space. We use a uniformly random choice among the following three transformations: (i) add or remove an instruction; instructions are added at a random position and have a random op and random arguments; to prevent programs from growing unnecessarily, instruction removal is twice as likely as addition; (ii) completely randomize all instructions in a component function by randomizing all their ops and arguments; or (iii) modify a randomly chosen argument of a randomly selected existing instruction. All categorical random choices are uniform. When modifying a real-valued constant, we multiply it by a uniform random number in $[0.5,2.0]$ and flip its sign with $10\%$ probability.

We upgrade the regularized evolution search method \cite{real2018regularized} to improve its performance in the following ways. These upgrades are justified empirically through ablation studies in Supplementary Section~\ref{ablation_comparison_thorough_sec}.

\textbf{Functional Equivalence Checking (FEC)}. The lack of heavy design of the search space allows for mutations that do not have an effect on the accuracy (\eg adding an instruction that writes to an address that is never read). When these mutations occur, the child algorithm behaves identically to its parent. To prevent these identically functioning algorithms from being repeatedly evaluated (\ie trained and validated in full many times), we keep an LRU cache mapping evaluated algorithm fingerprints to their accuracies. Before evaluating an algorithm, we first quickly fingerprint it and consult the cache to see if it has already been evaluated. If it has, we reuse the stored accuracy instead of computing it again. This way, we can keep the different implementations of the same algorithm for the sake of diversity: even though they produce the same accuracy now, they may behave differently upon further mutation.

To fingerprint an algorithm, we train it for 10 steps and validate it on 10 examples. The 20 resulting predictions are then truncated and hashed to produce an integer fingerprint. The cache holds $100\textrm{k}$ fingerprint--accuracy pairs.

\textbf{Parallelism}. In multi-process experiments, each process runs regularized evolution on its own population and the worker processes exchange algorithms through migration \cite{alba2002parallelism}. Every $100$--$10000$ evaluations, each worker uploads $50$ algorithms (half the population) to a central server. The server replies with $50$ algorithms sampled randomly across \textit{all} workers that are substituted into the worker's population.

\textbf{Dataset Diversity}. While the final evaluations are on binary CIFAR-10, in the experiments in Sections~\ref{main_result_sec} and~\ref{zoo_sec}, $50\%$ of the workers train and evaluate on binary MNIST instead of CIFAR-10. MNIST is a dataset of labeled hand-written digits \cite{lecun1998mnist}. We project MNIST to 256 dimensions in the same way we do for CIFAR-10. Supplementary Section~\ref{ablation_comparison_thorough_sec} demonstrates how searching on multiple MNIST-based and CIFAR-based tasks improves final performance on CIFAR-10, relative to searching only on multiple MNIST-based tasks or only on multiple CIFAR-based tasks.

\textbf{Hurdles}. We adopt the \textit{hurdles} upgrade to the evolutionary algorithm. This upgrade uses statistics of the population to early-stop the training of low performing models \cite{So2019TheET}. The early-stopping criterion is the failure to reach a minimum accuracy---the \emph{hurdle}. We alter the original implementation by setting the hurdle to the $75^{th}$ percentile of unique accuracies of the evolving population on a rolling basis (as opposed to the stationary value used in the original implementation). This alteration gives us more predictability over the resource savings: we consistently save $75\%$ of our compute, regardless of how the accuracy distribution shifts over the course of the search experiment.

\textbf{Terminating Degenerate Algorithms}. We terminate algorithms early if their calculations produce $\operatorname{NaN}$ or $\operatorname{Inf}$ values, and assign them a fixed minimum accuracy $a_{min}$ (we use $a_{min}\!\!=\!0$). Similarly, if an algorithm's error on any training example exceeds a threshold $e_{max}\!\!\gg\!\!1$ (we use $e_{max}\!\!=\!\!100$), we also stop that algorithm and assign it the accuracy $a_{min}$. Lastly, we time each algorithm as it is being executed and terminate it if its run-time exceeds a fixed threshold; we set this threshold to 4x the run-time of a plain neural network trained with gradient descent.

The experiment's meta-parameters (\eg $P$ and $T$) were either decided in smaller experiments (\eg $P$), taken from previous work (\eg $T$), or not tuned. Even when tuning parameters in smaller experiments, this was not done extensively (\eg no multi-parameter grid searches); typically, we tried a handful of values independently when each feature was introduced. For each experiment, we scaled---without tuning---some meta-parameters based on compute or hardware limitations. For example, compute-heavy tasks use smaller populations in order to save frequent checkpoints in case of machine reboots. Additional discrepancies between experiment configurations in the different sections are due to different researchers working independently.

}  

\supplsection{Task Generation Details}{task_details_sec}{

Sections~\ref{main_result_sec} and~\ref{zoo_sec} employ many binary classification tasks grouped into two sets, $\mathcal{T}_{search}$ and $\mathcal{T}_{select}$. We now describe how these tasks are generated. We construct a binary classification task by randomly selecting a pair of classes from CIFAR-10 to yield positive and negative examples. We then create a random projection matrix by drawing from a Gaussian distribution with zero mean and unit variance. We use the matrix to project the features of all the examples corresponding to the class pair to a lower dimension (\ie from the original 3072 to, for example, 16). The projected features are then standardized. This generates a proxy task that requires much less compute than the non-projected version. Each class pair and random projection matrix produce a different task. Since CIFAR-10 has 10 classes, there are 45 different pairs. For each pair we perform 100 different projections. This way we end up with 4500 tasks, each containing 8000/2000 training/validation examples. We use all the tasks from 36 of the pairs to form the $\mathcal{T}_{search}$ task set. The remaining tasks form $\mathcal{T}_{select}$.

}  

\supplsection{Detailed Search Experiment Setups}{detailed_experiment_setups_sec}{

Here we present details and method meta-parameters for experiments referenced in Section~\ref{results_sec}. These complement the ``Experiment Details'' paragraphs in the main text.

\textbf{Experiments in Section~\ref{difficulty_sec}, Figure~\ref{relative_difficulty_fig}}: Scalar/vector/matrix number of addresses: 4/3/1 (linear), 5/3/1 (affine). Fixed num.\ instructions for \setup/\predict/\learn: 5/1/4 (linear), 6/2/6 (affine). Expts.\ in this figure allow only minimal ops to discover a known algorithm, as follows. For ``linear backprop'' expts.: allowed \learn ops are \{OP3, OP4, OP19, OP24\}. For ``linear regressor'' expts.: allowed \setup ops are \{OP56, OP57\}, allowed predict ops are \{OP27\}, and allowed \learn ops are \{OP2, OP3, OP18, OP23\}. For ``affine backprop'' expts.: allowed \learn ops are \{OP1, OP2, OP3, OP18, OP23\}. For ``affine regressor'' expts.: allowed \setup ops are \{OP56, OP57\}, allowed \predict ops are \{OP1, OP27\}, and allowed \learn ops are \{OP1, OP2, OP3, OP18, OP23\}. 1 process, no server. Tasks: see \textit{Experiment Details} paragraph in main text. Evolution expts.: $P\!\!=\!\!1000$; $T\!\!=\!\!10$; $U\!\!=\!\!0.9$; we initialize the population with random programs; evals.\ per expt. for points in plot (left to right): $10\textrm{k}$, $10\textrm{k}$, $10\textrm{k}$, $100\textrm{k}$ (optimized for each problem difficulty to nearest factor of 10). Random search expts.: same num. memory addresses, same component function sizes, same total number of evaluations. These experiments are intended to be as simple as possible, so we do not use hurdles or additional data.

\textbf{Experiment in Section~\ref{difficulty_sec}, Figure~\ref{nn_fig}}: Scalar/vector/matrix number of addresses: 4/8/2. Fixed num.\ instructions for \setup/\predict/\learn: 21/3/9. In this figure, we only allow as ops those that appear in a two-layer neural network with gradient descent: allowed \setup ops are \{OP56, OP63, OP64\}, allowed \predict ops are \{OP27, OP31, OP48\}, and allowed \learn ops are \{OP2, OP3, OP16, OP18, OP23, OP25, OP28, OP40\}. Tasks: see \textit{Experiment Details} paragraph in main text. $P\!\!=\!\!1000$. $T\!\!=\!\!10$. $U\!\!=\!\!0.9$. $W\!\!=\!\!1\textrm{\normalfont{k}}$. Worker processes are uniformly divided into 4 groups, using parameters $T$/$D$/$P$ covering ranges in a log scale, as follows: $100\textrm{\normalfont{k}}$/$100$/$100$, $100\textrm{\normalfont{k}}$/$22$/$215$, $10\textrm{\normalfont{k}}$/$5$/$464$, and $100$/$1$/$1000$. Uses FEC. We initialize the population with random programs.

\textbf{Experiments in Section~\ref{main_result_sec}}: Scalar/vector/matrix number of addresses: 8/14/3. Maximum num.\ instructions for \setup/\predict/\learn: 21/21/45. All the initialization ops are now allowed for \setup: \{OP56, OP57, OP58, OP59, OP60, OP61, OP62, OP63, OP64\}. \predict and \learn use a longer list of 58 allowed ops: \{OP0, OP1, OP2, OP3, OP4, OP5, OP6, OP7, OP8, OP9, OP10, OP11, OP12, OP13, OP14, OP15, OP16, OP17, OP18, OP19, OP20, OP21, OP22, OP23, OP24, OP25, OP26, OP27, OP28, OP29, OP30, OP31, OP32, OP33, OP34, OP35, OP36, OP37, OP38, OP39, OP40, OP41, OP42, OP43, OP44, OP45, OP46, OP47, OP48, OP49, OP50, OP51, OP52, OP53, OP54, OP55, OP60, OP61\}---\emph{all} these ops are available to \emph{both} \predict and \learn. We use all the optimizations described in Section~\ref{discussion_sec}, incl.\ additional projected binary MNIST data. Worker processes are uniformly divided to perform each possible combination of tasks: $\{\textrm{\textit{projected binary CIFAR-10}}, \textrm{\textit{projected binary MNIST}}\}$ $\otimes$ $\{N\!\!=\!\!800\ \&\ E\!\!=\!\!1, N\!\!=\!\!8000\ \&\ E\!\!=\!\!1, N\!\!=\!\!800\ \&\ E\!\!=\!\!10\}$ $\otimes$ $\{D\!\!=\!\!1, D\!\!=\!\!10\}$ $\otimes$ $\{F\!\!=\!\!8, F\!\!=\!\!16, F\!\!=\!\!256\}$; where $N$ is the number of training examples, $E$ is the number of training epochs, and other quantities are defined in Section~\ref{methods_sec}. $P\!\!=\!\!100$. $T\!\!=\!\!10$. $U\!\!=\!\!0.9$. $W\!\!=\!\!10\textrm{\normalfont{k}}$ processes (commodity CPU cores). We initialize the population with empty programs.

\textbf{Experiments in Section~\ref{zoo_sec}}: Scalar/vector/matrix number of addresses: 10/16/4. Maximum num.\ instructions for \setup/\predict/\learn: 21/21/45. Allowed ops for \setup are \{OP56, OP57, OP58, OP59, OP60, OP61, OP62, OP63, OP64\}, allowed ops for \predict and \learn are \{OP0, OP1, OP2, OP3, OP4, OP5, OP6, OP7, OP8, OP9, OP10, OP11, OP12, OP13, OP14, OP15, OP16, OP17, OP18, OP19, OP20, OP21, OP22, OP23, OP24, OP25, OP26, OP27, OP28, OP29, OP30, OP31, OP32, OP33, OP34, OP35, OP36, OP37, OP38, OP39, OP40, OP41, OP42, OP43, OP44, OP45, OP46, OP47, OP48, OP49, OP50, OP51, OP52, OP53, OP54, OP55, OP63, OP64\}. These are the same ops as in the paragraph above, except for the minor accidental replacement of uniform for Gaussian initialization ops. We use FEC and hurdles. Workers use binary CIFAR-10 dataset projected to dimension 16. Half of the workers use $D\!\!=\!\!10$ (for faster evolution), and the other half use $D\!\!=\!\!100$ (for more accurate evaluation). $P\!\!=\!\!100$. $T\!\!=\!\!10$. $U\!\!=\!\!0.9$. Section~\ref{zoo_sec} considers three different task types: (1) In the ``few training examples'' task type (Figure~\ref{adapt_noisy_relu_fig}), experiments train each algorithm on 80 examples for 100 epochs for the experiments, while controls train on 800 examples for 100 epochs. (2) In the ``fast training'' task type (Figure~\ref{adapt_lr_decay_fig}), experiments train on 800 examples for 10 epochs, while controls train on 800 examples for 100 epochs. (3) In the ``multiple classes'' task type (Figure~\ref{adapt_norm_trick_fig}), experiments evaluate on projected 10-class CIFAR-10 
classification tasks, while controls evaluate on the projected binary CIFAR-10 classification tasks described before. The 10-class tasks are generated similarly to the binary tasks, as follows. Each task contains 45K/5K training/validation examples. Each example is a CIFAR-10 image projected to 16 dimensions using a random matrix drawn from a Gaussian distribution with zero mean and unit variance. This projection matrix remains fixed for all examples within a task. The data are standardized after the projection. We use 1000 different random projection matrices to create 1000 different tasks. $80\%$ of these tasks constitute $\mathcal{T}_{search}$ and the rest form $\mathcal{T}_{select}$. Since Section~\ref{main_result_sec} showed that we can discover reasonable models from scratch, in Section~\ref{zoo_sec}, we initialize the population with the simple two-layer neural network with gradient descent of Figure~\ref{nn_fig} in order to save compute.

}  

\supplsection{Evolved Algorithms}{full_evolved_algorithms_sec}{

In this section, we show the raw code for algorithms discovered by evolutionary search in Sections~\ref{main_result_sec} and~\ref{zoo_sec}. The code in those sections was simplified and therefore has superficial differences with the corresponding code here.

Supplementary Figure~\ref{raw_code_best_alg_fig} shows the raw code for the best evolved algorithm in Section~\ref{main_result_sec}. For comparison, Figure~\ref{semi_raw_code_best_alg_fig} shows the effect of removing redundant instructions through automated static analysis (details in Supplementary Section~\ref{interpretation_supp_sec}). For example, the instruction \textcode{v3 = gaussian(0.7,0.4)} has been deleted this way.

\begin{figure}[!ht]
\sbox1{\begin{subfigure}[t]{1.63in}
    \begin{code}{\textwidth}{3.9in}{codebackground}
    \codeline{\codedef{def} Setup():}
        \codeline{\codetab  s4 = uniform(0.6,0.2)}
        \codeline{\codetab  v3 = gaussian(0.7,0.4)}
        \codeline{\codetab  v12= gaussian(0.2,0.6)}
        \codeline{\codetab  s1 =}
        \codeline{\codetab\codetab\ uniform(-0.1,-0.2)}
    \codeskip
    \codeline{\codedef{def} Predict():}
        \codeline{\codetab v1 = v0 - v9 }
        \codeline{\codetab v5 = v0 + v9 }
        \codeline{\codetab v6 = dot(m1, v5) }
        \codeline{\codetab m1 = s2 * m2 }
        \codeline{\codetab s1 = dot(v6, v1)}
        \codeline{\codetab  s6 = cos(s4)}
    \codeskip
    \codeline{\codedef{def} Learn():}
        \codeline{\codetab s4 = s0 - s1}
        \codeline{\codetab s3 = abs(s1)}
        \codeline{\codetab m1 = outer(v1,v0)}
        \codeline{\codetab s5 = sin(s4)}
        \codeline{\codetab s2 = norm(m1) }
        \codeline{\codetab s7 = s5 / s2 }
        \codeline{\codetab s4 = s4 + s6}
        \codeline{\codetab v11= s7 * v1}
        \codeline{\codetab m1 = heaviside(m2)}
        \codeline{\codetab m1 = outer(v11,v5) } 
        \codeline{\codetab m0 = m1 + m0 }
        \codeline{\codetab v9 = uniform(2e-3,0.7) }
        \codeline{\codetab s7 = log(s0) }
        \codeline{\codetab s4 = std(m0) }
        \codeline{\codetab m2 = m2 + m0 }
        \codeline{\codetab m1 = s4 * m0 }
    \end{code}
    \caption{}
    \label{raw_code_best_alg_fig}
\end{subfigure}}
\sbox2{\begin{subfigure}[t]{1.57in}
    \begin{code}{\textwidth}{2.76in}{codebackground}
    \codeline{\codedef{def} Setup():}
    \codeskip
    \codeline{\codedef{def} Predict():}
        \codeline{\codetab v1 = v0 - v9 }
        \codeline{\codetab v5 = v0 + v9 }
        \codeline{\codetab v6 = dot(m1,v5) }
        \codeline{\codetab m1 = s2 * m2 }
        \codeline{\codetab s1 = dot(v6,v1)}
    \codeskip
    \codeline{\codedef{def} Learn():}
        \codeline{\codetab s4 = s0 - s1}
        \codeline{\codetab m1 = outer(v1,v0)}
        \codeline{\codetab s5 = sin(s4)}
        \codeline{\codetab s2 = norm(m1) }
        \codeline{\codetab s7 = s5 / s2 }
        \codeline{\codetab v11= s7 * v1}
        \codeline{\codetab m1 = outer(v11,v5) }
        \codeline{\codetab m0 = m1 + m0 }
        \codeline{\codetab v9 = }
        \codeline{\codetab\codetab\ uniform(2e-3,0.7) }
        \codeline{\codetab s4 = std(m0) }
        \codeline{\codetab m2 = m2 + m0 }
        \codeline{\codetab m1 = s4 * m0 }
    \end{code}
    \caption{}
    \label{semi_raw_code_best_alg_fig}
\end{subfigure}}
\setlength{\tabcolsep}{0pt}
\begin{tabular*}{\linewidth}{@{}@{\extracolsep{\fill}}cc}
\usebox1 & \usebox2\\
\end{tabular*}
\caption{(\subref{raw_code_best_alg_fig}) Raw code for the best evolved algorithm in Figure~\ref{experiment_progress_fig} (bottom--right corner) in Section~\ref{main_result_sec}. (\subref{semi_raw_code_best_alg_fig}) Same code after redundant instructions have been removed through static analysis.}
\end{figure}

Finally, the fully simplified version is in the bottom right corner of Figure~\ref{experiment_progress_fig} in the main text. To achieve this simplification, we used ablation studies to find instructions that can be removed or reordered. More details can be found in Supplementary Section~\ref{interpretation_supp_sec}. For example, in going from Supplementary Figure~\ref{semi_raw_code_best_alg_fig} to Figure~\ref{experiment_progress_fig}, we removed \textcode{s5 = sin(s4)} because its deletion does not significantly alter the accuracy. We also consistently renamed some variables (of course, this has no effect on the execution of the code).

Supplementary Figure~\ref{raw_code_adaptations_fig} shows the raw code for the algorithms in Figure~\ref{adapt_fig} in Section~\ref{zoo_sec}. Note that in Figure~\ref{adapt_fig}, we display a code snippet containing only a few selected instructions, while Supplementary Figure~\ref{raw_code_adaptations_fig} shows the programs in full.

\begin{figure*}[!htbp]
\sbox1{\begin{subfigure}[t]{2in}
    \begin{code}{\textwidth}{4in}{codebackground}
    \codeline{\codedef{def} Setup():}
    \codeline{\codetab s3 = 4.0e-3 }
    \codeskip
    \codeline{\codedef{def} Predict():}
    \codeline{\codetab v7 = v5 - v0 }
    \codeline{\codetab v3 = dot(m0, v0)}
    \codeline{\codetab s8 = s9 / s3}
    \codeline{\codetab v3 = v3 + v1}
    \codeline{\codetab m1 = heaviside(m3)}
    \codeline{\codetab v4 = maximum(v3, v7)}
    \codeline{\codetab s1 = dot(v4, v2)}
    \codeline{\codetab s4 = log(s9)}
    \codeline{\codetab v3 = bcast(s8)}
    \codeline{\codetab s7 = std(v1)}
    \codeline{\codetab v3 = v14 + v0}
    \codeline{\codetab m2 = matmul(m0, m2)}
    \codeskip
    \codeline{\codedef{def} Learn():}
    \codeline{\codetab v1 = gaussian(-0.50, 0.41)}
    \codeline{\codetab s4 = std(m0)}
    \codeline{\codetab s4 = s0 - s1}
    \codeline{\codetab v5 = gaussian(-0.48, 0.48)}
    \codeline{\codetab m1 = transpose(m3)}
    \codeline{\codetab s4 = s3 * s4}
    \codeline{\codetab v15 = v15 * v6}
    \codeline{\codetab v6 = s4 * v4}
    \codeline{\codetab v2 = v2 + v6}
    \codeline{\codetab v7 = s4 * v2}
    \codeline{\codetab v8 = heaviside(v3)}
    \codeline{\codetab v7 = v8 * v7}
    \codeline{\codetab m1 = outer(v7, v0)}
    \codeline{\codetab m0 = m0 + m1}
    \end{code}
    \caption{Raw code for the adaptation to few examples in Figure~\ref{adapt_noisy_relu_fig}.}
    \label{raw_code_noisy_relu_fig}
\end{subfigure}}
\sbox2{\begin{subfigure}[t]{2in}
    \begin{code}{\textwidth}{5.5in}{codebackground}
    \codeline{\codedef{def} Setup():}
        \codeline{\codetab s3 = 0.37 }
        \codeline{\codetab s1 = uniform(0.42, 0.66) }
        \codeline{\codetab s2 = 0.31 }
        \codeline{\codetab v13 = gaussian(0.69, 0.61) }
        \codeline{\codetab v1 = gaussian(-0.86, 0.97) }
    \codeskip
    \codeline{\codedef{def} Predict():}
        \codeline{\codetab m3 = m1 + m2 }
        \codeline{\codetab s6 = arccos(s0)}
        \codeline{\codetab v3 = dot(m0, v0)}
        \codeline{\codetab v3 = v3 - v0}
        \codeline{\codetab v11 = v2 + v9}
        \codeline{\codetab m2 = m0 - m2}
        \codeline{\codetab s4 = maximum(s8, s0)}
        \codeline{\codetab s7 = 1 / s6}
        \codeline{\codetab s6 = arctan(s0)}
        \codeline{\codetab s8 = minimum(s3, s1)}
        \codeline{\codetab v4 = maximum(v3, v10)}
        \codeline{\codetab s1 = dot(v4, v2)}
        \codeline{\codetab s1 = s1 + s2}
    \codeskip
    \codeline{\codedef{def} Learn():}
        \codeline{\codetab v1 = dot(m0, v5)}
        \codeline{\codetab s4 = s0 - s1}
        \codeline{\codetab s4 = s3 * s4}
        \codeline{\codetab s2 = s2 + s4}
        \codeline{\codetab m3 = matmul(m0, m3)}
        \codeline{\codetab m2 = bcast(v2, axis=0)}
        \codeline{\codetab v13 = s4 * v4}
        \codeline{\codetab v15 = v10 + v12}
        \codeline{\codetab v2 = v11 + v13}
        \codeline{\codetab v7 = s4 + v11}
        \codeline{\codetab v11 = v7 + v8}
        \codeline{\codetab s8 = s9 + s1}
        \codeline{\codetab m2 = m3 * m0}
        \codeline{\codetab s3 = arctan(s3)}
        \codeline{\codetab v8 = heaviside(v3)}
        \codeline{\codetab m2 = transpose(m1)}
        \codeline{\codetab s8 = heaviside(s6)}
        \codeline{\codetab s8 = norm(m3)}
        \codeline{\codetab v7 = v8 * v7}
        \codeline{\codetab m3 = outer(v7, v0)}
        \codeline{\codetab m0 = m0 + m3}
    \end{code}
    \caption{Raw code for the adaptation to fast training in Figure~\ref{adapt_lr_decay_fig}.}
    \label{raw_code_lr_decay_fig}
\end{subfigure}}
\sbox3{\begin{subfigure}[t]{2in}
    \begin{code}{\textwidth}{7.1in}{codebackground}
    \codeline{\codedef{def} Setup():}
        \codeline{\codetab m3 = uniform(0.05, 0.11)}
        \codeline{\codetab s1 = uniform(0.31, 0.90)}
        \codeline{\codetab v18 = uniform(-0.49, 4.41)}
        \codeline{\codetab s1 = -0.65}
        \codeline{\codetab m5 = uniform(0.21, 0.22)}
        \codeline{\codetab v9 = gaussian(0.64, 7.8e-3)}
        \codeline{\codetab s1 = -0.84}
    \codeskip
    \codeline{\codedef{def} Predict():}
        \codeline{\codetab s1 = abs(s1)}
        \codeline{\codetab v15 = norm(m1, axis=1)}
        \codeline{\codetab v15 = dot(m0, v0)}
        \codeline{\codetab v8 = v19 - v0}
        \codeline{\codetab v15 = v8 + v15}
        \codeline{\codetab v7 = max(v1, v15)}
        \codeline{\codetab v13 = min(v5, v4)}
        \codeline{\codetab m2 = transpose(m2)}
        \codeline{\codetab v10 = v13 * v0}
        \codeline{\codetab m7 = heaviside(m3)}
        \codeline{\codetab m4 = transpose(m7)}
        \codeline{\codetab v2 = dot(m1, v7)}
        \codeline{\codetab v6 = max(v2, v9)}
        \codeline{\codetab v2 = v2 + v13}
        \codeline{\codetab v11 = heaviside(v17)}
        \codeline{\codetab s1 = sin(s1)}
        \codeline{\codetab m3 = m6 - m5}
        \codeline{\codetab v19 = heaviside(v14)}
        \codeline{\codetab v10 = min(v12, v7)}
    \codeskip
    \codeline{\codedef{def} Learn():}
        \codeline{\codetab m5 = abs(m7)}
        \codeline{\codetab v8 = v1 - v2}
        \codeline{\codetab m2 = transpose(m2)}
        \codeline{\codetab v8 = s1 * v8}
        \codeline{\codetab v15 = v15 - v4}
        \codeline{\codetab v4 = v4 + v8}
        \codeline{\codetab s1 = arcsin(s1)}
        \codeline{\codetab v15 = mean(m3, axis=1)}
        \codeline{\codetab v12 = v11 * v11}
        \codeline{\codetab m4 = heaviside(m5)}
        \codeline{\codetab m6 = outer(v8, v7)}
        \codeline{\codetab s1 = sin(s1)}
        \codeline{\codetab s1 = exp(s0)}
        \codeline{\codetab m1 = m1 + m3}
        \codeline{\codetab m5 = outer(v15, v6)}
        \codeline{\codetab m2 = transpose(m1)}
        \codeline{\codetab s1 = exp(s0)}
        \codeline{\codetab v12 = uniform(0.30, 0.33)}
        \codeline{\codetab s1 = minimum(s0, s1)}
        \codeline{\codetab m5 = m5 * m7}
        \codeline{\codetab v9 = dot(m2, v8)}
        \codeline{\codetab v9 = v10 * v9}
        \codeline{\codetab v3 = norm(m7, axis=1)}
        \codeline{\codetab s1 = mean(m1)}
        \codeline{\codetab m2 = outer(v9, v0)}
        \codeline{\codetab m0 = m0 + m2}
    \end{code}
    \caption{Raw code for the adaptation to multiple classes in Figure~\ref{adapt_norm_trick_fig}.}
    \label{raw_code_norm_trick_fig}
\end{subfigure}}
\begin{tabular*}{\linewidth}{@{}@{\extracolsep{\fill}}ccc}
\usebox1 & \usebox2 & \usebox3\\
\end{tabular*}
\caption{Raw evolved code for algorithm snippets in Figure~\ref{adapt_fig} in Section~\ref{zoo_sec}.}
\label{raw_code_adaptations_fig}
\end{figure*}

}  

\supplsection{Algorithm Selection and Evaluation}{reevaluation_supp_sec}{

We first run search experiments evaluating algorithms on the projected binary classification tasks sampled from $\mathcal{T}_{search}$ and collect the best performing candidate from each experiment. The measure of performance is the median accuracy across tasks. Then, we rank these candidates by evaluating them on tasks sampled from $\mathcal{T}_{select}$ and we select the highest-ranking candidate (this is analogous to typical model selection practice using a validation set). The highest ranking algorithm is finally evaluated on the binary classification tasks using CIFAR-10 data with the original dimensionality (3072). 

Because the algorithms are initially evolved on tasks with low dimensionality (16) and finally evaluated on the full-size dimensionality (3072), their hyperparameters must be tuned on the full-size dimensionality before that final evaluation. To do this, we treat all the constants in the algorithms as hyperparameters and jointly tune them using random search. For each random search trial, each constant is scaled up or down by a random factor sampled between $0.001$ and $1000$ on a log-scale. We allowed up to $10\mathrm{k}$ trials to tune the hyperparameters, but only a few hundred were required to tune the best algorithm in Figure~\ref{experiment_progress_fig}---note that this algorithm only has 3 constants. To make comparisons with baselines fair, we tune these baselines using the same amount of resources that went into tunining \emph{and evolving} our algorithms. All hyperparameter-tuning trials use 8000 training and 2000 validation examples from the CIFAR-10 \emph{training} set. After tuning, we finally run the tuned algorithms on 2000 examples from the held-out CIFAR-10 \emph{test} set. We repeat this final evaluation with 5 different random seeds and report the mean and standard deviation. We stress that the CIFAR-10 test set was used only in this final evaluation, and never in $\mathcal{T}_{search}$ or $\mathcal{T}_{select}$.

In our experiments, we found a \textit{hyperparameter coupling} phenomenon that hinders algorithm selection and tuning. ML algorithms usually make use of hyperparameters (\eg learning rate) that need to be tuned for different datasets (for example, when the datasets have very different input dimensions or numbers of examples). Similarly, the evolved algorithms also contain hyperparameters that need to be adjusted for different datasets. If the hyperparameters are represented as constants in the evolved algorithm, we can identify and tune them on the new dataset by using random search. However, it is harder to tune them if a hyperparameter is instead computed from other variables. For example, in some evolved algorithms, the learning rate $s_2$ was computed as $s_2 = norm(v_1)$ because the best value for $s_2$ coincides with the L2-norm of $v_1$ on $\mathcal{T}_{search}$. However, when we move to a new dataset with higher dimensions, the L2-norm of $v_1$ might no longer be a good learning rate. This can cause the evolved algorithms' performance to drop dramatically on the new dataset. To resolve this, we identify these parameters by manual inspection of the evolved code. We then manually decouple them: in the example, we would set $s_2$ to a constant that we can tune with random-search. This recovers the performance. Automating the decoupling process would be a useful direction for future work.

}  

\supplsection{Interpreting Algorithms}{interpretation_supp_sec}{

It is nontrivial to interpret the raw evolved code and decide which sections of it are important. We use the following procedures to help with the interpretation of discovered algorithms:

(a) We clean up the raw code (\eg Figure~\ref{raw_code_best_alg_fig}) by automatically simplifying programs. To do this, we remove redundant instructions through static analysis, resulting in code like that in Figure~\ref{semi_raw_code_best_alg_fig}. Namely, we analyze the computations that lead to the final prediction and remove instructions that have no effect. For example, we remove instructions that initialize variables that are never used.

(b) We focus our attention on code sections that reappear in many independent search experiments. This is a sign that such code sections may be beneficial. For example, Section~\ref{zoo_sec} applied this procedure to identify adaptations to different tasks.

(c) Once we have hypotheses about interesting code sections, we perform ablations/\emph{knock-outs}, where we remove the code section from the algorithm to see if there is a significant loss in accuracy. As an example, for  Section~\ref{main_result_sec}, we identified 6 interesting code sections in the best evolved algorithm to perform ablations. For each ablation, we removed the relevant code section, then tuned all the hyperparameters / constants again, and then computed the loss in validation accuracy. 4 out of the 6 ablations caused a large drop in accuracy. These are the ones that we discussed in Section~\ref{main_result_sec}. Namely, (1) the addition of noise to the input ($-0.16\%$); (2) the bilinear model ($-1.46\%$); (3) the normalized gradients ($-1.20\%$); and (4) the weight averaging ($-4.11\%$). The remaining 2 code sections show no significant loss upon ablation, and so were removed for code readability.
Also for readability, we reorder some instructions when this makes no difference to the accuracy either (\eg we move related code lines closer to each other). After this procedure, the code looks like that in Figure~\ref{experiment_progress_fig}. 

(d) If an ablation suggests that a code section is indeed helpful to the original algorithm, we then perform a \emph{knock-in}. That is, we insert the code section into simpler algorithms to see if it improves their performance too. This way we confirmed the usefulness of the 4 code sections mentioned in (c), for example.

}  

\supplsection{More Search Method Ablations}{ablation_comparison_thorough_sec}{

To verify the effectiveness of the upgrades mentioned in Supplementary Section~\ref{detailed_methods_optimization_sec}, we conduct ablation studies using the experiment setup of Section~\ref{main_result_sec}, except for the following simplifications to reduce compute: (i) we limit the ops to only those that are necessary to construct a neural network trained with gradient descent (as was done for Figure~\ref{nn_fig}), \ie allowed \setup ops are \{OP57, OP64, OP65\}, allowed \predict ops are \{OP28, OP32, OP49\}, and allowed \learn ops are \{OP3, OP4, OP17, OP19, OP24, OP26, OP29, OP41\}; (ii) we reduce the projected dimensionality from 256 to 16, and (iii) we use $1\textrm{k}$ processes for 5 days. Additionally, the ablation experiments we present use $T=8000,E=10$ for all tasks. This slight difference is not intentionally introduced, but rather is a product of our having studied our method before running our experiments in the main text; we later on found that using more epochs did not change the conclusions of the studies or improve the results.

Supplementary Tables~\ref{methods_ablation_large_compute_table},~\ref{methods_ablation_mid_compute_table}, and~\ref{methods_ablation_low_compute_table} display the results. Note that Figure~\ref{ablation_comparison_fig} presents a subset of this data in plot form (the indexes 1--4 along the horizontal axis labels in that figure coincide with the ``Index'' column in this table). We find that all four upgrades are beneficial across the three different compute scales tested.

\begin{table*}[!ht]
\caption{Ablation studies. Each row summarizes the results of 30 search runs under one given experimental setting. Rows \#0--4 all use the same setting, except for the search method: each row implements an upgrade to the method and shows the resulting improvement. ``Best Accuracy'' is the accuracy of the best algorithm for each experiment ($\pm 2\operatorname{SEM}$), evaluated on unseen projected binary CIFAR-10 tasks. ``Success Fraction'' is the fraction ($\pm 2\sigma$) of those experiments that produce algorithms that are more accurate than a plain neural network trained with gradient descent ($0.750$). This fraction helps us estimate the likelihood of high performing outliers, which we are keenly interested in. The experimental setting for row \#5 is the same as row \#3, except that instead of using both projected binary CIFAR-10 and projected binary MNIST data for the search, we use only projected binary MNIST data (as for other rows, the accuracy is reported on projected binary CIFAR-10 data). Row \#5 indicates that searching \emph{completely} on MNIST data is not as helpful as searching partially on it (row \#3). Overall, rows \#0--4 suggest that all four upgrades are beneficial.}
\label{methods_ablation_large_compute_table}
\begin{center}
\begin{small}
\begin{sc}
\begin{tabular}{clcc}
\toprule
Index & Description & Best Accuracy & Success Fraction \\
\midrule
0 & Baseline & $0.703 \pm 0.002$ & $0.00 \pm 0.00$  \\
1 & + Migration & $0.707 \pm 0.004$ & $0.00 \pm 0.00$ \\
2 & + Functional Equivalence Check & $0.724 \pm 0.006$ & $0.13 \pm 0.12$ \\
3 & + 50\% MNIST Data & $0.729 \pm 0.008$ & $0.27 \pm 0.16$  \\
4 & + Hurdles & $0.738 \pm 0.008$ & $0.53 \pm 0.18$ \\
\\
5 & Experiment 3 w/ 100\% MNIST Data & $0.720 \pm 0.003$ & $0.00 \pm 0.00$ \\
\bottomrule
\end{tabular}
\end{sc}
\end{small}
\end{center}
\end{table*}

\begin{table*}[!ht]
\caption{This is the same as Supplementary Table~\ref{methods_ablation_large_compute_table}, except at a lower compute scale (100 processes). Each setup was run 100 times. The results are similar to those with more compute and support the same conclusions. Thus, the observed benefits are not specific to a single compute scale.}
\label{methods_ablation_mid_compute_table}
\begin{center}
\begin{small}
\begin{sc}
\begin{tabular}{clcc}
\toprule
Index & Description & Best Accuracy & Success Fraction \\
\midrule
0 & Baseline & $0.700 \pm 0.002$ & $0.00 \pm 0.00$ \\
1 & + Migration & $0.704 \pm 0.000$ & $0.00 \pm 0.00$ \\
2 & + Functional Equivalence Check & $0.706 \pm 0.001$ & $0.00 \pm 0.00$ \\
3 & + 50\% MNIST Data & $0.710 \pm 0.002$ & $0.02 \pm 0.03$ \\
4& + Hurdles & $0.714 \pm 0.003$ & $0.10 \pm 0.06$ \\
\\
5 & Experiment 3 w/ 100\% MNIST Data & $0.700 \pm 0.004$ & $0.00 \pm 0.00$ \\
\bottomrule
\end{tabular}
\end{sc}
\end{small}
\end{center}
\end{table*}

\begin{table*}[!ht]
\caption{This is the same as Supplementary Tables~\ref{methods_ablation_large_compute_table} and~\ref{methods_ablation_mid_compute_table}, except at an even lower compute scale (10 processes). Each setup was run 100 times. The results are consistent with those with more compute, but we no longer observe successes (``successes'' defined in Table~\ref{methods_ablation_large_compute_table}).}
\label{methods_ablation_low_compute_table}
\begin{center}
\begin{small}
\begin{sc}
\begin{tabular}{clcc}
\toprule
Index & Description & Best Accuracy & Success Fraction \\
\midrule
0 & Baseline & $0.700 \pm 0.001$ & $0.00 \pm 0.00$ \\
1 & + Migration & $0.701 \pm 0.001$ & $0.00 \pm 0.00$ \\
2 & + Functional Equivalence Check & $0.702 \pm 0.001$ & $0.00 \pm 0.00$ \\
3 & + 50\% MNIST Data & $0.704 \pm 0.001$ & $0.00 \pm 0.00$ \\
4& + Hurdles & $0.705 \pm 0.001$ & $0.00 \pm 0.00$ \\
\\
5 & Experiment 3 w/ 100\% MNIST Data & $0.694 \pm 0.002$ & $0.00 \pm 0.00$ \\
\bottomrule
\end{tabular}
\end{sc}
\end{small}
\end{center}
\end{table*}

\pagebreak{}

}  

\supplsection{Baselines}{baselines_sec}{

The focus of this study was not the search method but we believe there is much room for future work in this regard. To facilitate comparisons with other search algorithms on the same search space, in this section we provide convenient baselines at three different compute scales.

All baselines use the same setting, a simplified version of that in Section~\ref{main_result_sec}, designed to use less compute. In particular, here we severely restrict the search space to be able to reach results quickly. Scalar/vector/matrix number of addresses: 5/9/2. Maximum num.\ instructions for \setup/\predict/\learn: 7/11/23. Allowed \setup ops: \{OP57, OP60, OP61, OP62, OP63, OP64, OP65\}, allowed \predict ops: \{OP2, OP24, OP28, OP32, OP49\}, allowed \learn ops: \{OP2, OP3, OP4, OP17, OP19, OP24, OP26, OP29, OP41\}. Experiments end after each process has run 100B training steps---\ie the training loop described in Section~\ref{methods_search_space_sec} runs 100B times. (We chose ``training steps'' instead of ``number of algorithms'' as the experiment-ending criterion because the latter varies due to early stopping, FEC, \etc. We also did not choose ``time'' as the experiment-ending criterion to make comparisons hardware-agnostic. For reference, each experiment took roughly 12 hours on our hardware.) $P\!\!=\!\!100$; $T\!\!=\!\!10$, $U\!\!=\!\!0.9$. All workers evaluate on the same projected binary CIFAR-10 tasks as in Section~\ref{main_result_sec}, except that we project to 16 dimensions instead of 256. Each search evaluation is on 10 tasks and the numbers we present here are evaluations on $\mathcal{T}_{select}$ using 100 tasks. We initialize the population with empty programs.

The results of performing 100 repeats of these experiments at three different compute scales are summarized in Table~\ref{external_baselines_table}; note, each process is run on a single commodity CPU core. We additionally compare our full search method from Section~\ref{main_result_sec}, labeled ``Full'', and a more ``Basic'' search setup, which does not use FEC, hurdles, or MNIST data.

\begin{table*}[!ht]
\caption{Baselines on the simplified setting with the restricted search space (see Supplementary Section~\ref{baselines_sec} for details). ``Best Accuracy'' is the best evaluated accuracy on unseen projected binary classification tasks for each run ($\pm 2\operatorname{SEM}$), ``Linear Success Fraction'' is the fraction ($\pm 2\sigma$) of those accuracies that are above the evaluated accuracy of logistic regression trained with gradient descent ($0.702$), and ``NN Success Fraction'' is the fraction ($\pm 2\sigma$) of those accuracies that are above the evaluated accuracy of a plain neural network trained with gradient descent ($0.729$). Using success fractions as a metric  helps us estimate the likelihood of discovering high performing outliers, which we are keenly interested in. Each experiment setup was run 100 times. The ``Full'' method is the one we used in Section~\ref{main_result_sec}; the ``Basic'' method is the same, but with no FEC, no hurdles, and no MNIST data.}
\label{external_baselines_table}
\begin{center}
\begin{small}
\begin{sc}
\begin{tabular}{ccccc}
\toprule
Method & Number of Processes & Best Accuracy & Linear Success Fraction & NN Success Fraction \\
\midrule
Basic & 1 & $0.671 \pm 0.004$ &  $0.01 \pm 0.02$ &  $0.00 \pm 0.00$ \\
Basic & 10 & $0.681 \pm 0.005$  &  $0.07 \pm 0.05$ &  $0.00 \pm 0.00$ \\
Basic & 100 & $0.691 \pm 0.004$  & $0.26 \pm 0.09$ & $0.00 \pm 0.00$ \\
\\
Full & 1 & $0.684 \pm 0.003$ &  $0.03 \pm 0.03$ &  $0.00 \pm 0.00$ \\
Full & 10 & $0.693 \pm 0.003$ &  $0.23 \pm 0.08$ &  $0.03 \pm 0.03$ \\
Full & 100 & $0.707 \pm 0.003$ & $0.59 \pm 0.10$ & $0.11 \pm 0.06$ \\
\bottomrule
\end{tabular}
\end{sc}
\end{small}
\end{center}
\end{table*}

}  

\end{document}